\documentclass[journal]{IEEEtai}

\usepackage[colorlinks,urlcolor=blue,linkcolor=blue,citecolor=blue]{hyperref}
\usepackage{pgfplots}  
\pgfplotsset{width=6.6cm,compat=1.7}  
\usepackage{color,array}
\usepackage[most]{tcolorbox}
\usepackage{subcaption}
\usepackage[utf8]{inputenc}
\usepackage{graphicx}
\usepackage{amsmath}
\usepackage{todonotes}
\usepackage{amsthm}
\usepackage{xcolor}
\usepackage{ulem}
\usepackage{booktabs}
\usepackage{algorithm}
\usepackage{algorithmic}
\usepackage{graphicx}
\usepackage{amsmath}
\usepackage{amsmath,amssymb,amsfonts}
\usepackage{algorithm}
\usepackage{booktabs}
\usepackage{algorithmic}
\usepackage{bm}
\usepackage{multirow}
\usepackage{tabulary,xspace}
\usepackage{amssymb}
\usepackage{color}
\usepackage{multirow}
\usepackage{amssymb}
\usepackage{mathtools}
\usepackage{makecell}
\let\labelindent\relax
\usepackage{enumitem}
\usepackage{caption}
\usepackage{pgf-pie}
\newcolumntype{L}{>{\raggedright\arraybackslash}X}
\usepackage{ltablex}
\usepackage{siunitx}
\usepackage{caption}
\usepackage{lipsum}
\usepackage{amsmath}
\usepackage{graphicx}
\usepackage[colorlinks, linkcolor=blue, urlcolor=blue]{hyperref}
\usepackage{pifont}

\usepackage{tikz}

\begin{document}

\title{OD-VIRAT: A Large-Scale Benchmark for Object Detection in Realistic Surveillance Environments} 

\author{Hayat Ullah, Abbas Khan, Arslan Munir \IEEEmembership{Member, IEEE}, and Hari Kalva \IEEEmembership{Senior Member, IEEE}



\thanks{Hayat Ullah, Abbas Khan, and Arlsan Munir are with the Intelligent Systems, Computer Architecture, Analytics, and Security Laboratory (ISCAAS Lab), Department of Electrical Engineering and Computer Science, Florida Atlantic University, Boca Raton, Florida 33431, USA. (e-mail: hullah2024@fau.edu, abbaskhan2024@fau.edu, arslanm@fau.edu)}
\thanks{Hari Kalva is with the Multimedia Lab, Department of Electrical Engineering and Computer Science, Florida Atlantic University, Boca Raton, Florida 33431, USA. (e-mail: hkalva@fau.edu).}

}


\maketitle

\begin{abstract}
Realistic human surveillance datasets are crucial for training and evaluating computer vision models under real-world conditions, facilitating the development of robust algorithms for human and human-interacting object detection in complex environments. These datasets need to offer diverse and challenging data to enable a comprehensive assessment of model performance and the creation of more reliable surveillance systems for public safety. To this end, we present two visual object detection benchmarks named OD-VIRAT Large and OD-VIRAT Tiny, aiming at advancing visual understanding tasks in surveillance imagery. The video sequences in both benchmarks cover 10 different scenes of human surveillance recorded from significant height and distance. The proposed benchmarks offer rich annotations of bounding boxes and categories, where OD-VIRAT Large has 8.7 million annotated instances in 599,996 images and OD-VIRAT Tiny has 288,901 annotated instances in 19,860 images. This work also focuses on benchmarking state-of-the-art object detection architectures, including RETMDET, YOLOX, RetinaNet, DETR, and Deformable-DETR on this object detection-specific variant of VIRAT dataset. To the best of our knowledge, it is the first work to examine the performance of these recently published state-of-the-art object detection architectures on realistic surveillance imagery under challenging conditions such as complex backgrounds, occluded objects, and small-scale objects. The proposed benchmarking and experimental settings will help in providing insights concerning the performance of selected object detection models and set the base for developing more efficient and robust object detection architectures. Our code and models are released at: \href{https://github.com/iscaas/AFOSR-HAR-2021-2025/tree/main/OD-VIRAT}{\textcolor{purple}{\texttt{https://github.com/iscaas/OD-VIRAT}}}
\end{abstract}
\begin{IEEEImpStatement}
Realistic surveillance data is essential for state-of-the-art object detection models as it ensures accurate and reliable performance in real-world applications. The availability of realistic surveillance data can significantly benefit the research community by providing a robust foundation for developing, testing, and refining advanced object detection algorithms, thereby accelerating innovation and improving model generalization across diverse real-world scenarios. To this end,  we introduce two variants of OD-VIRAT, named OD-VIRAT Large and OD-VIRAT Tiny, both aimed at advancing object detection tasks in surveillance imagery. Besides, we also provide the benchmarking of newly introduced dataset by conducting extensive experiments using the state-of-the-art object detection architectures, thereby providing valuable insights into the performance of the selected models and set the baseline for the development of increasingly efficient and resilient architectures in this field 
\end{IEEEImpStatement}


\section{Introduction}

\begin{figure}[t]
  \centering
  \begin{subfigure}{0.23\textwidth}
    \includegraphics[width=\linewidth]{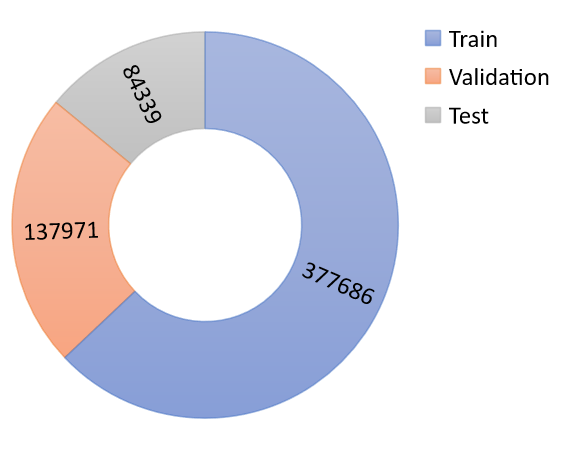}
    \caption{OD-VIRAT Large}
    \label{fig:first}
  \end{subfigure}
  \begin{subfigure}{0.23\textwidth}
    \includegraphics[width=\linewidth]{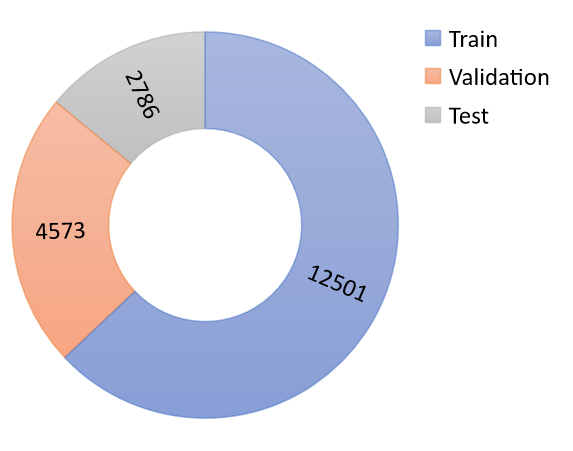}
    \caption{OD-VIRAT Tiny}
    \label{fig:second}
  \end{subfigure}
  \caption{Graphical illustration of train, validation, and test split of \textbf{OD-VIRAT Large} and \textbf{OD-VIRAT Tiny} dataset.}
  \label{fig:od_virat_split}
\end{figure}

\IEEEPARstart{O}{bject} detection in surveillance environments plays a pivotal role in ensuring the safety and security of various settings, ranging from public spaces to private facilities. This sophisticated technology leverages advanced computer vision algorithms to automatically identify and track objects of interest within a video feed or image stream. By efficiently detecting and analyzing objects such as people, vehicles, and suspicious items, surveillance systems equipped with object detection algorithms empower security personnel to promptly respond to potential threats and incidents, thereby enhancing overall situational awareness and proactive risk management.\\
\begin{figure}[t]
	\centering
	\includegraphics[width=\linewidth]{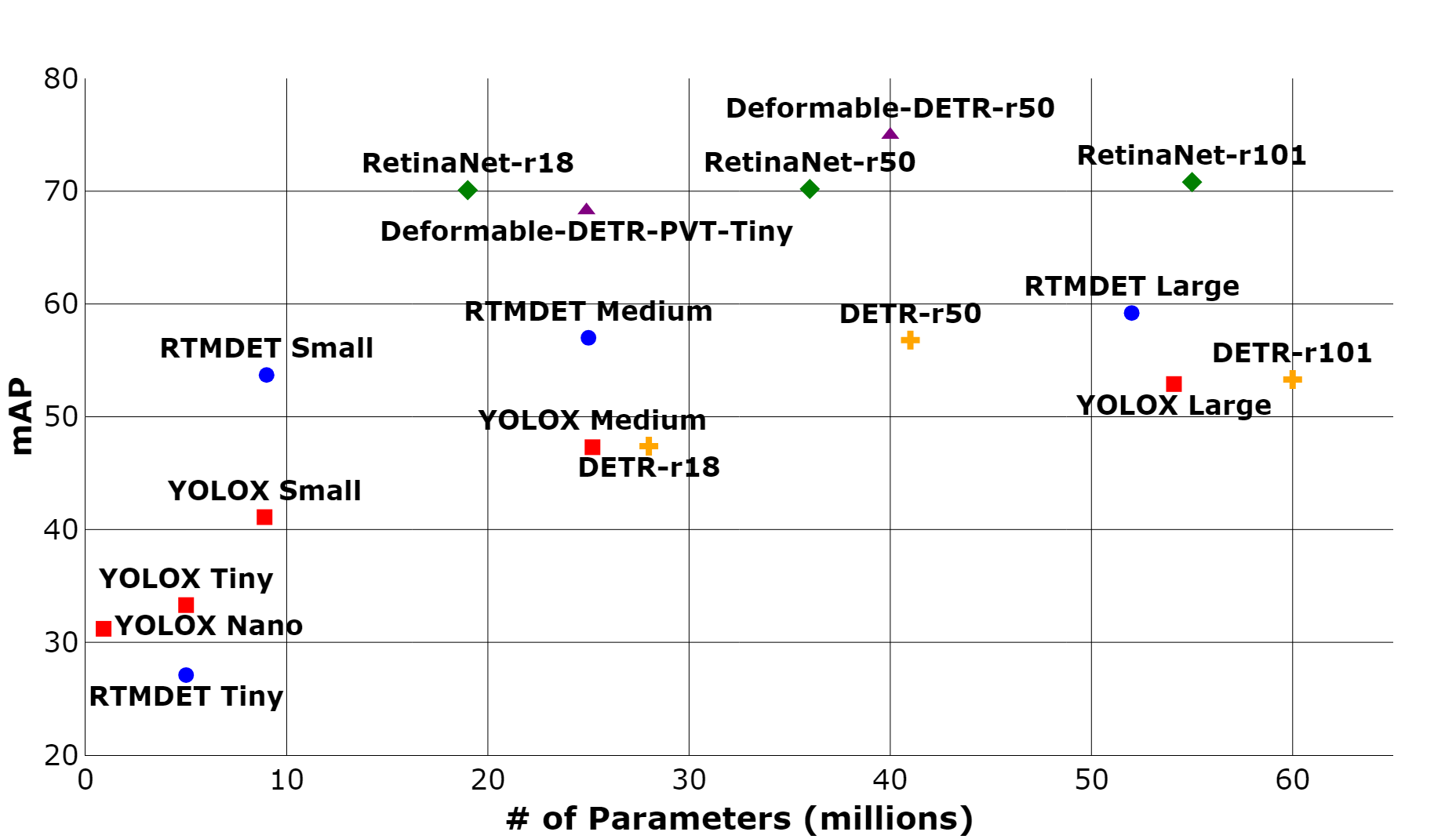}
	\caption{\textit{Model Complexity vs Accuracy (mAP)} trade-off comparison: We evaluate the performance of five main-stream object detection architectures on \textbf{OD-VIRAT Tiny} dataset and compared the obtained mAP values against model complexities (\# of parameters). The Deformable-DETR architecture with resnet50 backbone outperform other counterparts by obtaining the best mAP value.}
	\label{fig:perfor_vs_param}
\end{figure}
\indent In recent years, the field of object detection has witnessed remarkable advancements, fueled by the rapid growth of high-resolution cameras, a large amount of available data, powerful computing resources, and the rapid evolution of deep learning methodologies \cite{cheng2022anchor, chen2023diffusiondet, pu2024rank}. These innovations have enabled surveillance systems to achieve unprecedented levels of precision \cite{roy2022fast,li2022dual}, scalability, and real-time processing capabilities \cite{li2022yolov6,wang2023yolov7}. From crowded urban areas to critical infrastructure sites, the deployment of object detection technology has become indispensable for law enforcement agencies, transportation authorities, retail establishments, and other stakeholders tasked with safeguarding public safety and protecting valuable assets. As the demand for robust surveillance solutions continues to grow, ongoing research efforts strive to enhance the performance, efficiency, and adaptability of object detection algorithms \cite{chen2023diffusiondet,li2022exploring,cheng2022anchor} to meet the evolving needs of modern security environments. However, despite the significant advancements, realistic surveillance object detection datasets are often unavailable. Existing datasets \cite{lin2014microsoft, everingham2010pascal} lack the complexities of surveillance environments, hindering algorithm development and evaluation. Bridging this gap is essential to ensure that object detection technology meets the demands of modern security settings.\\
\indent The necessity of realistic surveillance object detection datasets is essential in the efforts to further advance the capabilities of surveillance systems. These datasets serve as the foundation upon which machine learning models are trained, tested, and refined to accurately identify objects in complex real-world scenarios. By encompassing diverse environmental conditions, lighting variations, occlusions, and object poses, realistic datasets enable researchers and developers to create more robust and reliable object detection algorithms. Consequently, the creation and availability of comprehensive and representative surveillance datasets are essential to drive innovation, benchmark performance, and ultimately, enhance the effectiveness of object detection algorithms for practical and real-time applications.\\
\indent To this end, we introduce two object detection benchmarks, OD-VIRAT Large and OD-VIRAT Tiny, to enhance visual comprehension tasks in surveillance imagery. These benchmarks comprise video sequences capturing 10 distinct scenes of human surveillance, recorded from considerable height and distance. The proposed benchmarks offer rich annotations of bounding boxes and categories, where OD-VIRAT Large has 8.7 million annotated instances in 599,996 images (comprising of three splits, i.e., [train, validation, test] = [377686, 137971, 84339]) and OD-VIRAT Tiny has 288,901 annotated instances in 19,860 images (comprising of three splits, i.e., [train, validation, test] = [12501, 4573, 2786]), as depicted in Fig.~\ref{fig:od_virat_split}.\\
\indent More precisely, the key contributions of this work are as follows:
\begin{itemize}
    \item Two realistic surveillance object detection datasets are contributed in this paper: OD-VIRAT Large and OD-VIRAT Tiny datasets. OD-VIRAT Large provides 8.7 million annotated instances in 599,996 images depicting 10 different surveillance scenes. OD-VIRAT Tiny has 288,901 annotated instances in 19,860 images collected from 10 different surveillance scenes. The images featured in both benchmarks present challenging characteristics, including complex backgrounds, occluded tiny objects, and objects of varying sizes, which make their detection difficult.
    \item To the best of our knowledge, this work presents the first empirical evaluation study of five state-of-the-art object detection architectures on challenging surveillance imagery prepared for object detection tasks as depicted in Fig.~\ref{fig:perfor_vs_param}. We believe that the reported empirical evaluation in this paper will set the standard baseline for future object detection architectures to further enhance the object detection performance on challenging surveillance images.
    \item To evaluate the effectiveness of object detection architectures in uncertain environments, we extensively investigate the performance of models under the scope of our experimental settings on images perturbed with different types of noises/artifacts, including Gaussian Noise, Motion Blur, Snow, and Elastic Transform.
\end{itemize}

\indent The remainder of this paper is structured as follows. Section~\ref{sec:od_virat} presents the detailed overview of OD-VIRAT Large and Tiny datasets, discussing the extension process followed by statistical details and annotations across both benchmarks. In Section~\ref{sec:benchmarking_method}, we present the benchmarking methods for OD-VIRAT dataset, comprising the utilization of state-of-the-art object detection architectures. Next, Section~\ref{sec:experimetnal_results} covers the detailed experimental settings and obtained quantitative and qualitative results on OD-VIRAT Tiny dataset. Finally, Section~\ref{sec:conclusion} concludes this paper.
\begin{table*}[t]
\caption{Statistical comparison of our \textbf{OD-VIRAT Large} and \textbf{OD-VIRAT Tiny} with commonly used object detection benchmarks, including BDD100K, PASCAL VOC 2007, PASCAL VOC 2012, COCO 2017, Objects365, Open Images V7, VisDrone, and V3Det datasets.}
\centering
\resizebox{1\linewidth}{!}{
\begin{tabular}{c|c|c|c|c|c|c|c|c|c|c}
    \toprule
    \multirow{2}{*}{Dataset} & \multicolumn{3}{c|}{\# of Scenes} & \multicolumn{3}{c|}{\# of Videos} & \multicolumn{3}{c|}{\# of Images} & \multirow{2}{*}{Resoultion}\\ 
    \cline{2-4} \cline{5-7} \cline{8-10}
    &train&validation&test&train&validation&test&train&validation&test&\\
    \hline
    BDD100K \cite{yu2020bdd100k} &---&---&---&---&---&---&70,000&10,000&20,000&{1280$\times$720}\\ \hline
    PASCAL VOC 2007 \cite{everingham2010pascal} &---&---&---&---&---&---&2,501&2,510&---&varying resolution\\ \hline
    PASCAL VOC 2012 \cite{voc2017} &---&---&---&---&---&---&5,717&5,823&---&varying resolution\\ \hline
    COCO 2017 \cite{coco2017} &---&---&---&---&---&---&1,18,287&5,000&40,670&varying resolution\\ \hline
    Objects365 \cite{shao2019objects365} &---&---&---&---&---&---& 6,00,000& 38,000& 1,00,000 & varying resolution \\ \hline
    Open Images V7 \cite{OpenImages} &---&---&---&---&---&---&9,011,219 &41,620 &125,436 &varying resolution \\ \hline
    VisDrone \cite{9573394} &---&---&---&---&---&---&6,471 &548 &3,190 & 2000 $\times$ 1500 \\ \hline
    V3Det \cite{wang2023v3det} &---&---&---&---&---&---&1,83,354 &29,821 &29,863 & 660 $\times$ 789\\ \hline
    OD-VIRAT Tiny&10&10&8&156&52&52&12,501&4,573&2,786&(1920$\times$1080),(1280$\times$720) \\\hline
    OD-VIRAT Large&10&10&8&156&52&52&3,77,686&1,37,971&84,339&(1920$\times$1080),(1280$\times$720)\\
    \bottomrule
\end{tabular}}
\label{tab:statis}
\end{table*}
\section{OD-VIRAT Benchmark Dataset}
\label{sec:od_virat}
The proposed OD-VIRAT dataset is the object detection variant of VIRAT Ground 2.0 dataset \cite{oh2011large}. Originally, VIRAT Ground 2.0 dataset is developed for continuous visual event recognition tasks, comprising 329 surveillance videos captured through stationary ground cameras mounted at significant heights (mostly at the top of buildings) across 11 different scenes. The recorded scenes include construction sites, parking lots, streets, and open outdoor areas. The authors of VIRAT Ground 2.0 dataset used different models of HD video cameras to capture scenes at different resolutions (1920$\times$1080 and 1280$\times$720) and frame rates range (25$\thicksim$30 FPS). For annotation, they created three different annotation files for each video namely, {\fontfamily{qcr}\selectfont events.txt}, {\fontfamily{qcr}\selectfont objects.txt}, and {\fontfamily{qcr}\selectfont mapping.txt}. The {\fontfamily{qcr}\selectfont events.txt} file contains temporal annotation of events, the {\fontfamily{qcr}\selectfont objects.txt} file contains the bounding box coordinates of moving objects, and the {\fontfamily{qcr}\selectfont mapping.txt} file provides the mapping between bounding box annotations and event annotations to form complete annotation for a single video.\\
\begin{figure}[t]
	\centering
	\includegraphics[width=\linewidth]{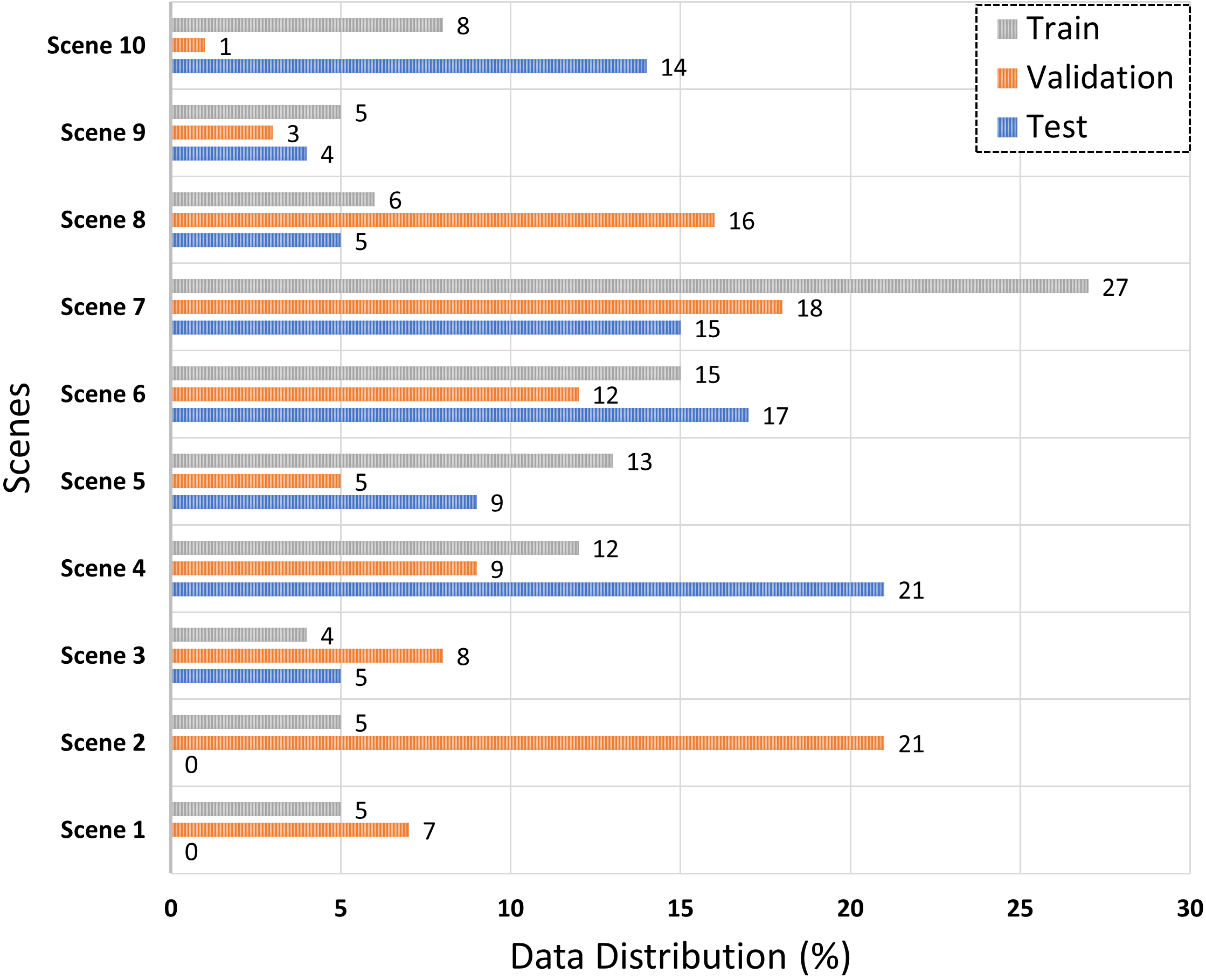}
	\caption{Data distribution across 10 scenes in train, validation, and tests of \textbf{OD-VIRAT Large} dataset.}
	\label{fig:od_virat_large_stats}
\end{figure}
\indent Our motivation behind introducing OD-VIRAT dataset is based on two key aspects: (1) the existence\footnote{\scriptsize Videos are selected from \href{https://viratdata.org/}{VIRAT Ground 2.0} dataset based on the existence of objects relevant to object-level annotations} of high-quality surveillance videos, and (2) the availability\footnote{\scriptsize The bounding box coordinates are retrieved from the {\fontfamily{qcr}\selectfont objects.txt} file in VIRAT Ground 2.0 dataset.} of object-level annotations. Based on the relevance between videos and their corresponding object annotations, we selected 260 videos out of 329 videos from VIRAT Ground 2.0 dataset. The remaining 69 videos are excluded as per their incorrect temporal relation between objects and their annotations. The retrieved 260 videos contain visuals of 10 distinct scenes, which are subsequently divided into three sets: train, validation, and test set. The train set contains 156 videos whereas the validation and test sets contain 52 videos each, respectively. It is worth mentioning here that, the train and validation sets consist of 10 scenes each, while the test set contains 8 scenes. As per requirement of the object detection task, we converted each set (including training, validation, and test) of videos into frames with two different frame-skip strategies that include \textit{0 frame-skip} and \textit{30 frame-skip} strategy. The \textit{0 frame-skip} strategy is used to create OD-VIRAT Large dataset, whereas \textit{30 frame-skip} is adopted to create OD-VIRAT Tiny dataset. The detailed statistical overview of OD‑VIRAT Large and OD‑VIRAT Tiny, together with comparisons to the BDD100K \cite{yu2020bdd100k}, PASCAL VOC 2007 \cite{everingham2010pascal}, PASCAL VOC 2012 \cite{voc2017}, COCO 2017 \cite{coco2017}, Objects365 \cite{shao2019objects365}, Open Images V7 \cite{OpenImages}, VisDrone \cite{9573394}, and V3Det \cite{wang2023v3det} datasets, is presented in Table~\ref{tab:statis}.
\begin{figure}[t]
	\centering
	\includegraphics[width=\linewidth]{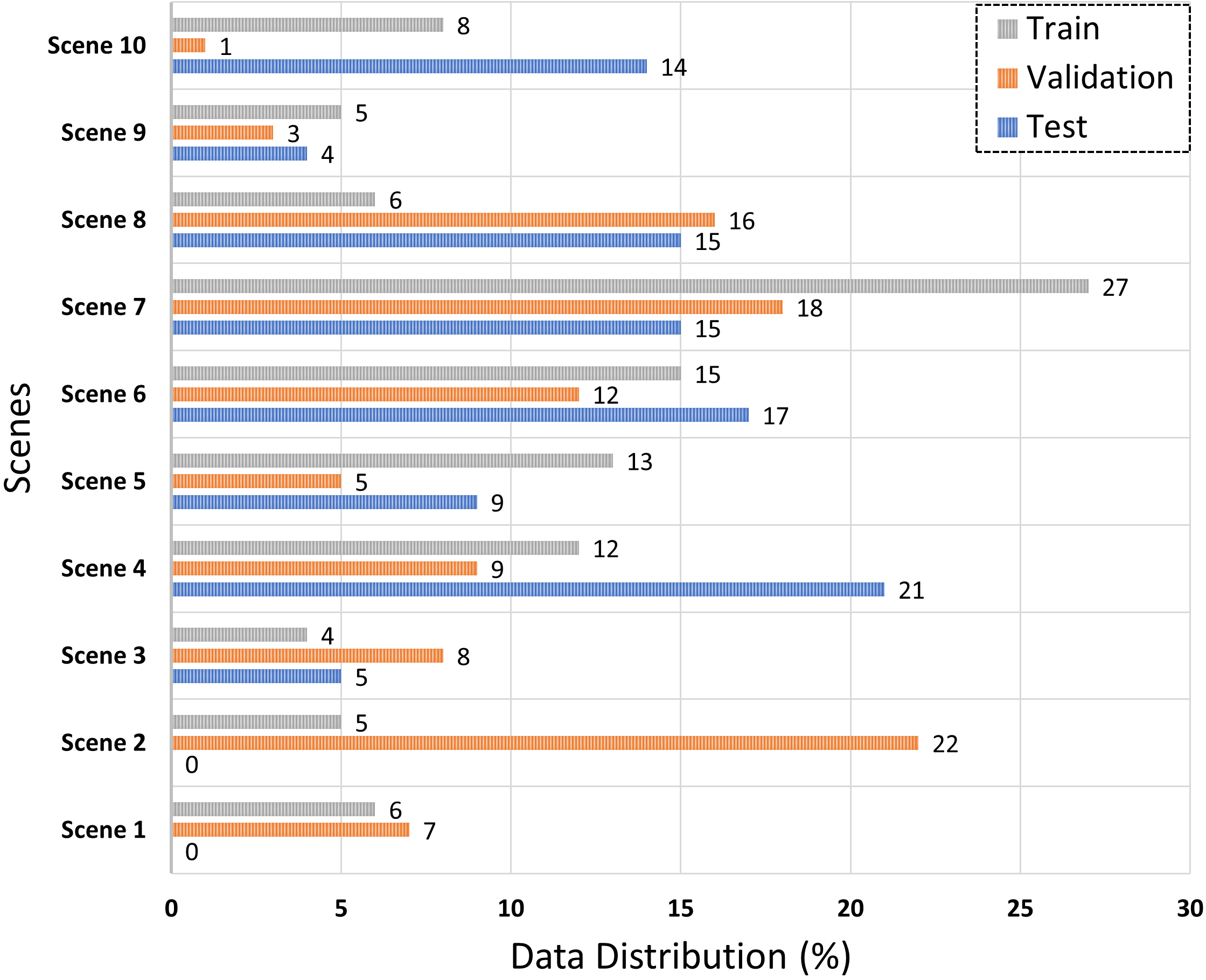}
	\caption{Data distribution across 10 scenes in train, validation, and tests of \textbf{OD-VIRAT Tiny} dataset.}
	\label{fig:od_virat_tiny_stats}
\end{figure}
\subsection{OD-VIRAT Large}
\label{sec:od_virat_large}
The OD-VIRAT Large dataset is created by converting videos into frames from each set without skipping frames (i.e., \textit{0 frame-skip} strategy). The goal of this OD-VIRAT Large dataset is to provide a gigantic object detection benchmark dataset to computer vision research community. It contains 10 different scenes captured in the daytime, including the surveillance of five objects of interest that are (1) {\fontfamily{qcr}\selectfont Bike/Bicycle}, (2) {\fontfamily{qcr}\selectfont Car}, (3) {\fontfamily{qcr}\selectfont Carrying\_object}, (4) {\fontfamily{qcr}\selectfont Person}, and (5) {\fontfamily{qcr}\selectfont Vehicle}. All scenes are recorded from a significant height and distance, covering a large area of the scene under the observation. The detailed statistical overview of OD-VIRAT Large dataset can be found in Table~\ref{tab:statis}, listing the number of scenes, videos, and frames for training, validation, and testing. Overall, the total number of frames in OD-VIRAT Large dataset is 599,996, after splitting the total number of train, validation, and test frames are 377686 (62.94\%), 137971 (23\%), and 84339 (14.05\%), respectively. Moreover, the percentage of frames per scene distributed across train, validation, and test sets is depicted in Fig.~\ref{fig:od_virat_large_stats}.
\subsection{OD-VIRAT Tiny}
\label{sec:od_virat_tiny}
The OD-VIRAT Tiny dataset is the smaller version of OD-VIRAT Large dataset, generated by converting videos into frames using \textit{30 frame-skip} strategy. This tiny version contains considerably less data in comparison with OD-VIRAT Large, yet covers all scenes and number of classes that are available in the larger version of OD-VIRAT. The detailed statistical overview of OD-VIRAT Tiny dataset is presented in Table~\ref{tab:statis}, providing number total number of scenes, videos, and frames across train, validation, and test sets. The total number of frames in OD-VIRAT Tiny is 19,860, where the number of frames in train, validation, and test splits are 12501 (62.94\%), 4573 (23.03\%), and 2786 (14.02\%), respectively. It is worth mentioning here that, the OD-VIRAT Tiny dataset is approximately $30\times$ smaller than OD-VIRAT Large dataset. Similarly, the reduction ratio of train, validation, and test sets in OD-VIRAT Tiny are approximately $30\times$, $30\times$, and $30\times$, respectively. Moreover, the distribution percentage of frames per scene across train, validation, and test sets is illustrated in Fig.~\ref{fig:od_virat_tiny_stats}.

\subsection{Annotations}
\label{sec:annotations}
\begin{figure}[t]
	\centering
	\begin{subfigure}{0.23\linewidth}
		\includegraphics[width=\linewidth]{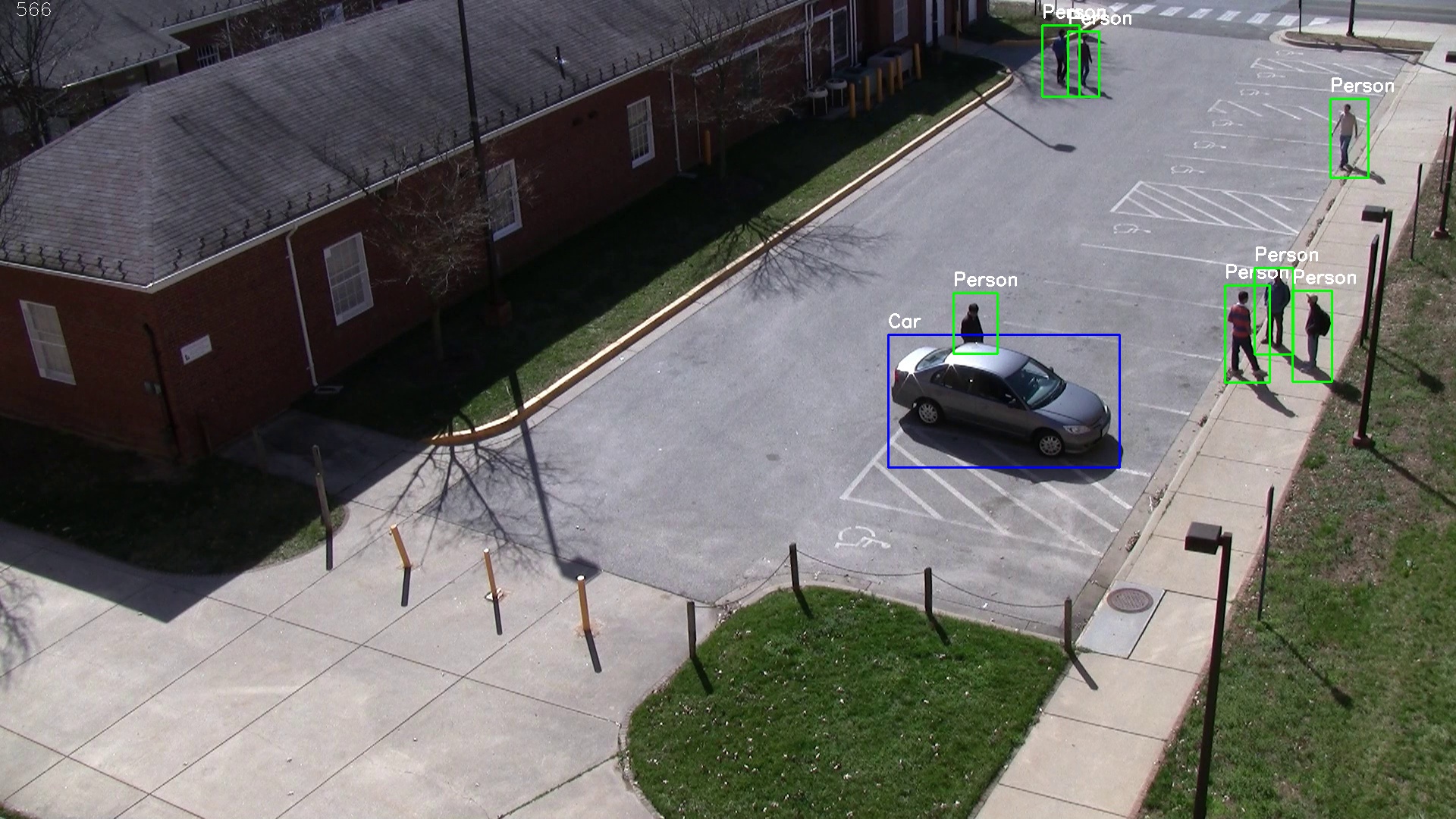}
		\caption{Scene 1}
		\label{fig:subfigA}
	\end{subfigure}
	\begin{subfigure}{0.23\linewidth}
		\includegraphics[width=\linewidth]{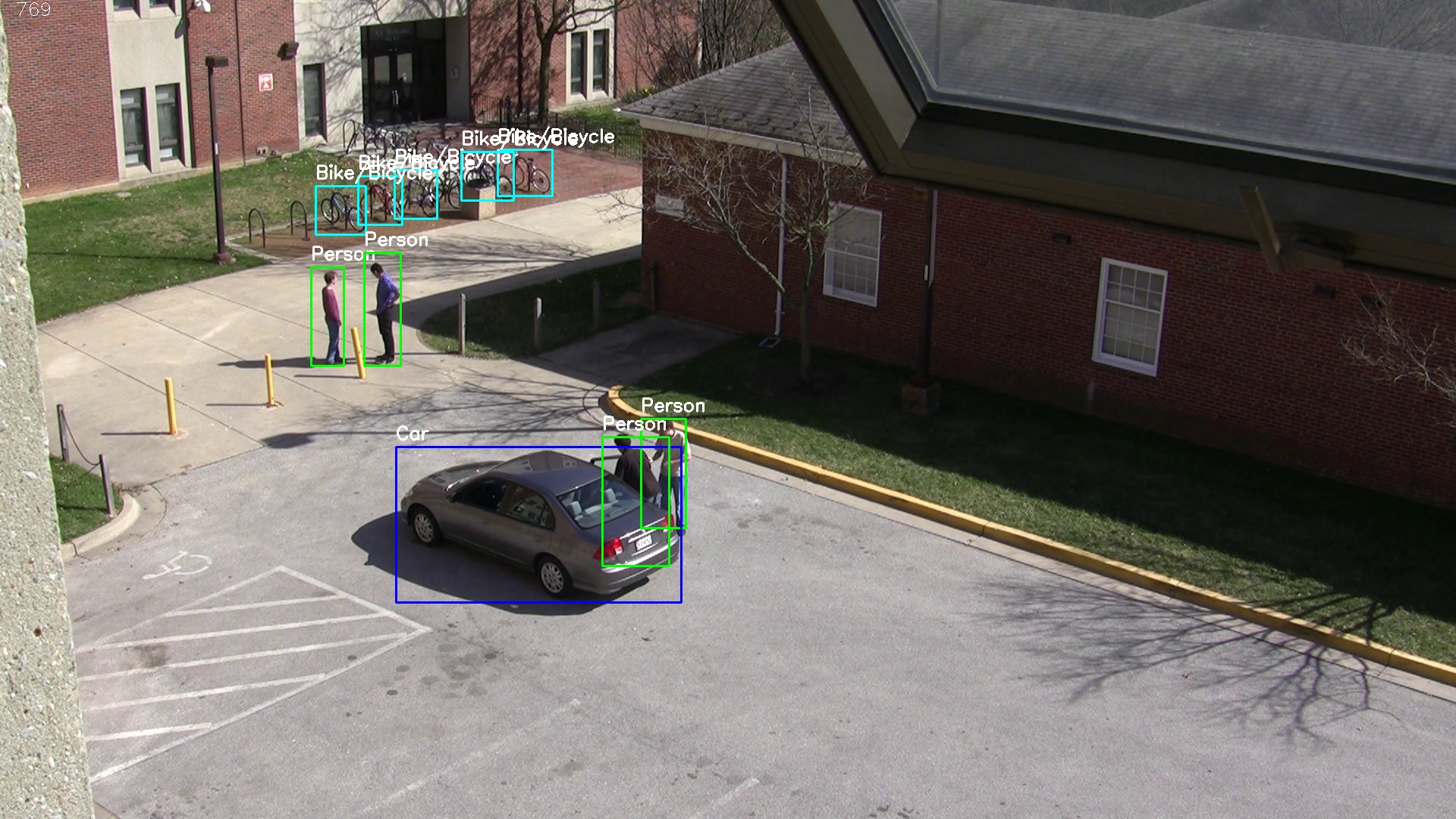}
		\caption{Scene 2}
		\label{fig:subfigB}
	\end{subfigure}
	\begin{subfigure}{0.23\linewidth}
	        \includegraphics[width=\linewidth]{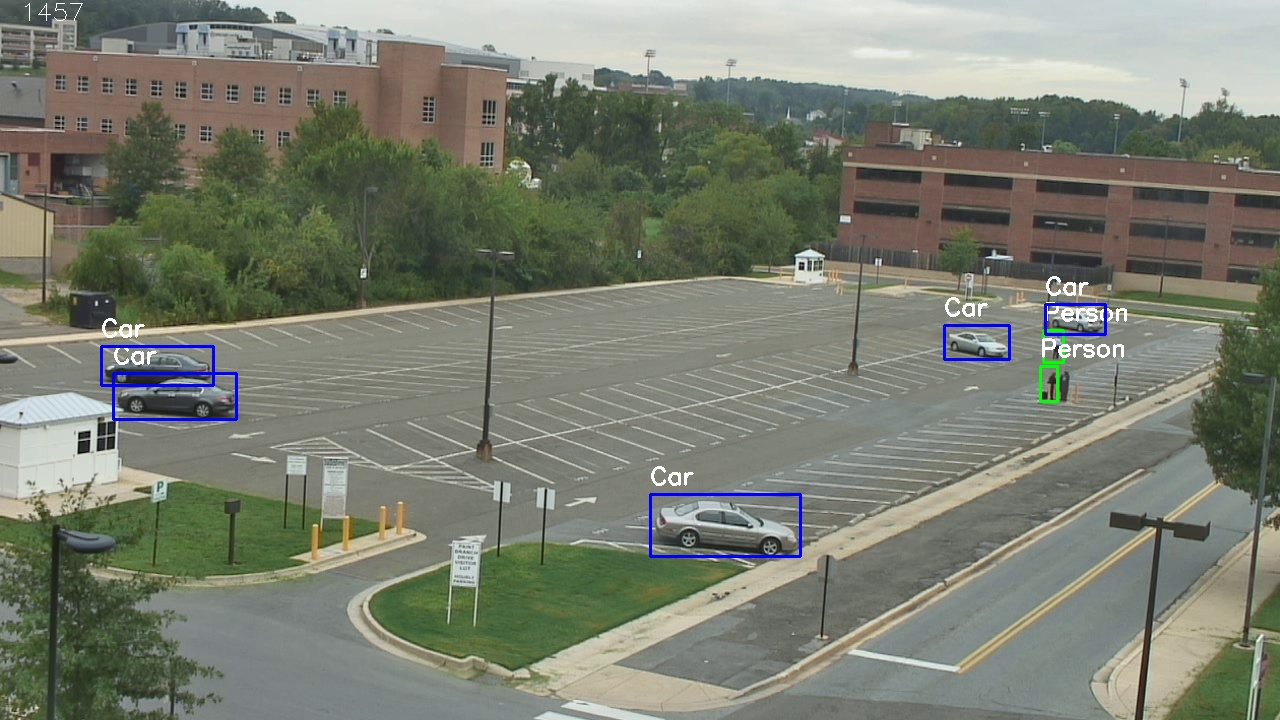}
	        \caption{Scene 3}
	        \label{fig:subfigC}
    \end{subfigure}
    \begin{subfigure}{0.23\linewidth}
	        \includegraphics[width=\linewidth]{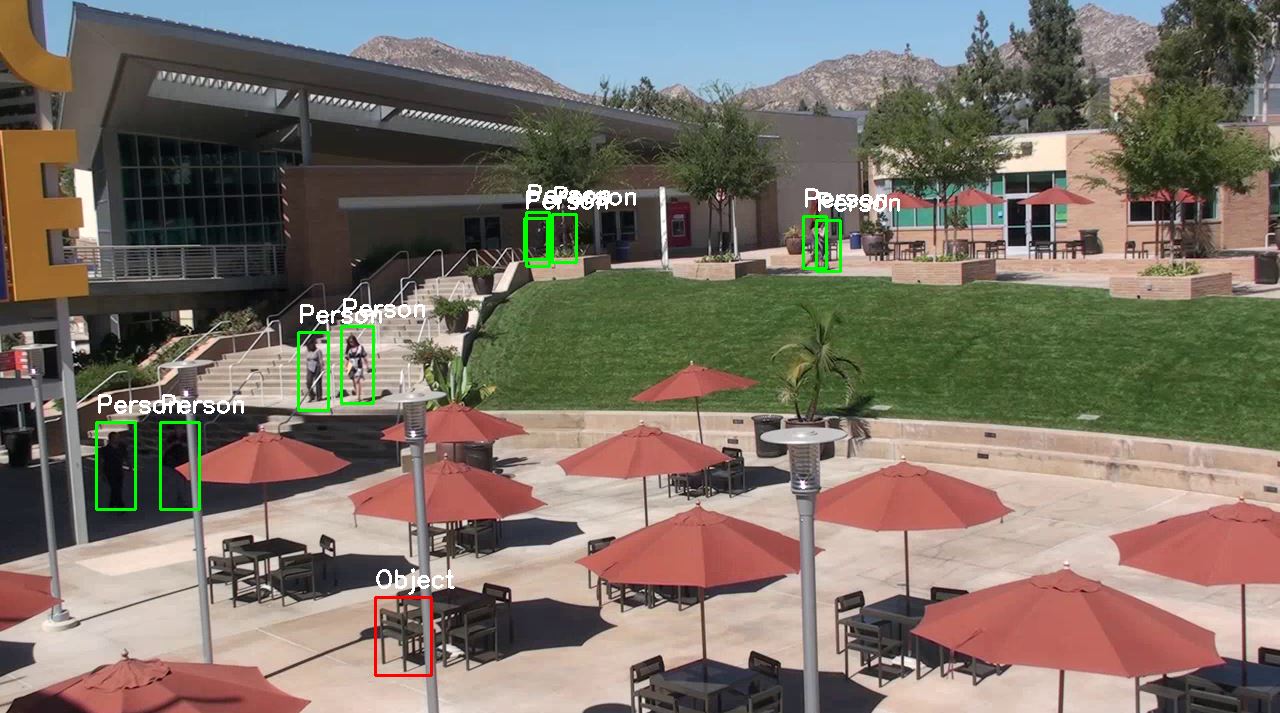}
	        \caption{Scene 4}
	        \label{fig:subfigC}
    \end{subfigure}
    \begin{subfigure}{0.23\linewidth}
	        \includegraphics[width=\linewidth]{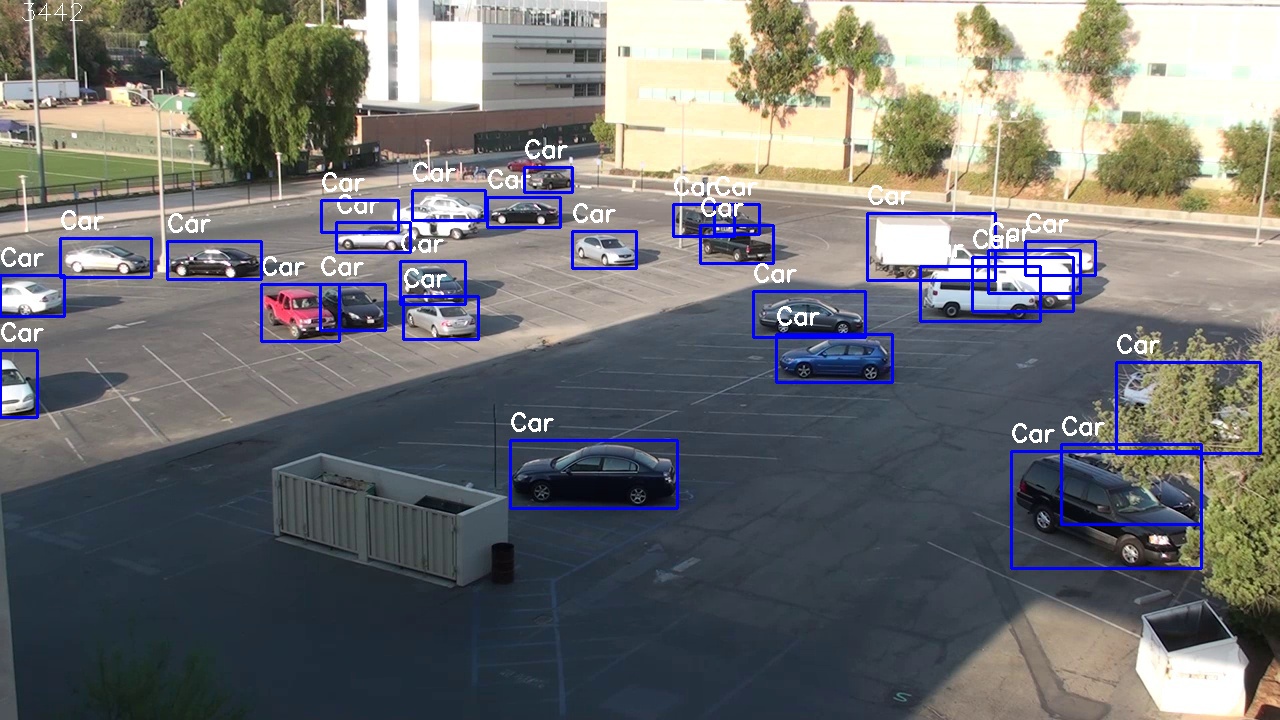}
	        \caption{Scene 5}
	        \label{fig:subfigC}
    \end{subfigure}
    \begin{subfigure}{0.23\linewidth}
	        \includegraphics[width=\linewidth]{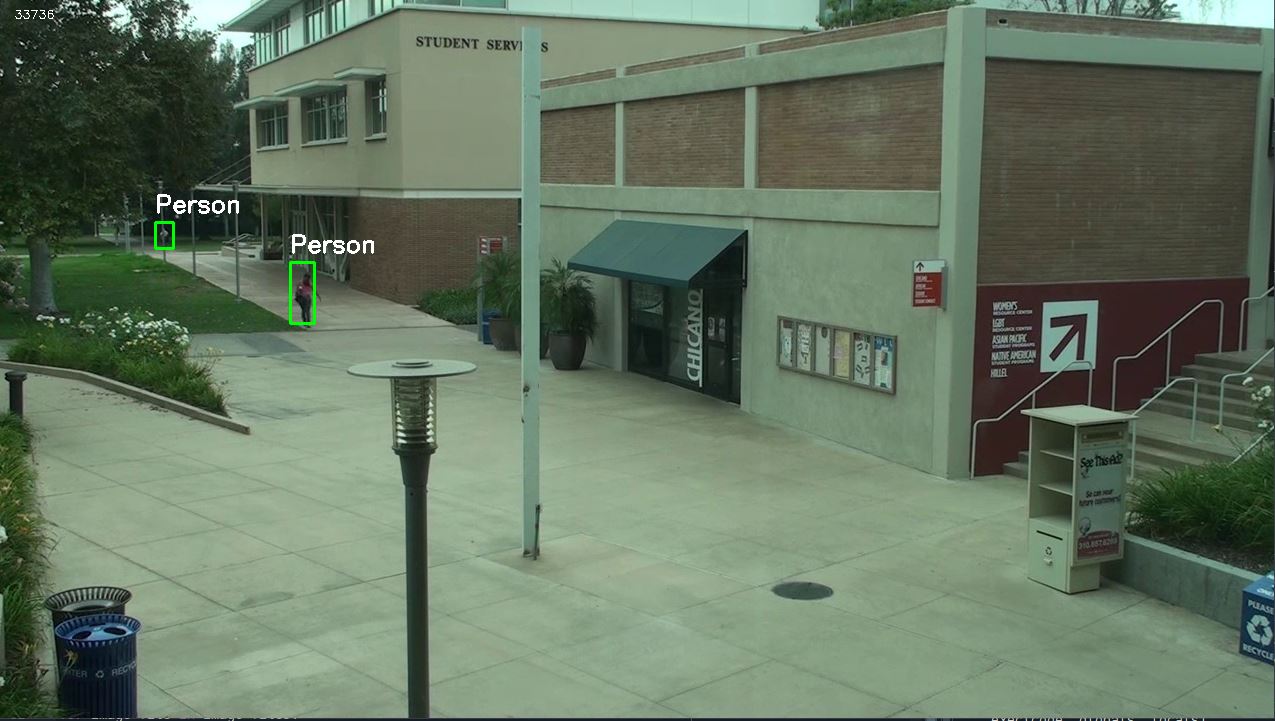}
	        \caption{Scene 6}
	        \label{fig:subfigC}
    \end{subfigure}
	\begin{subfigure}{0.23\linewidth}
	        \includegraphics[width=\linewidth]{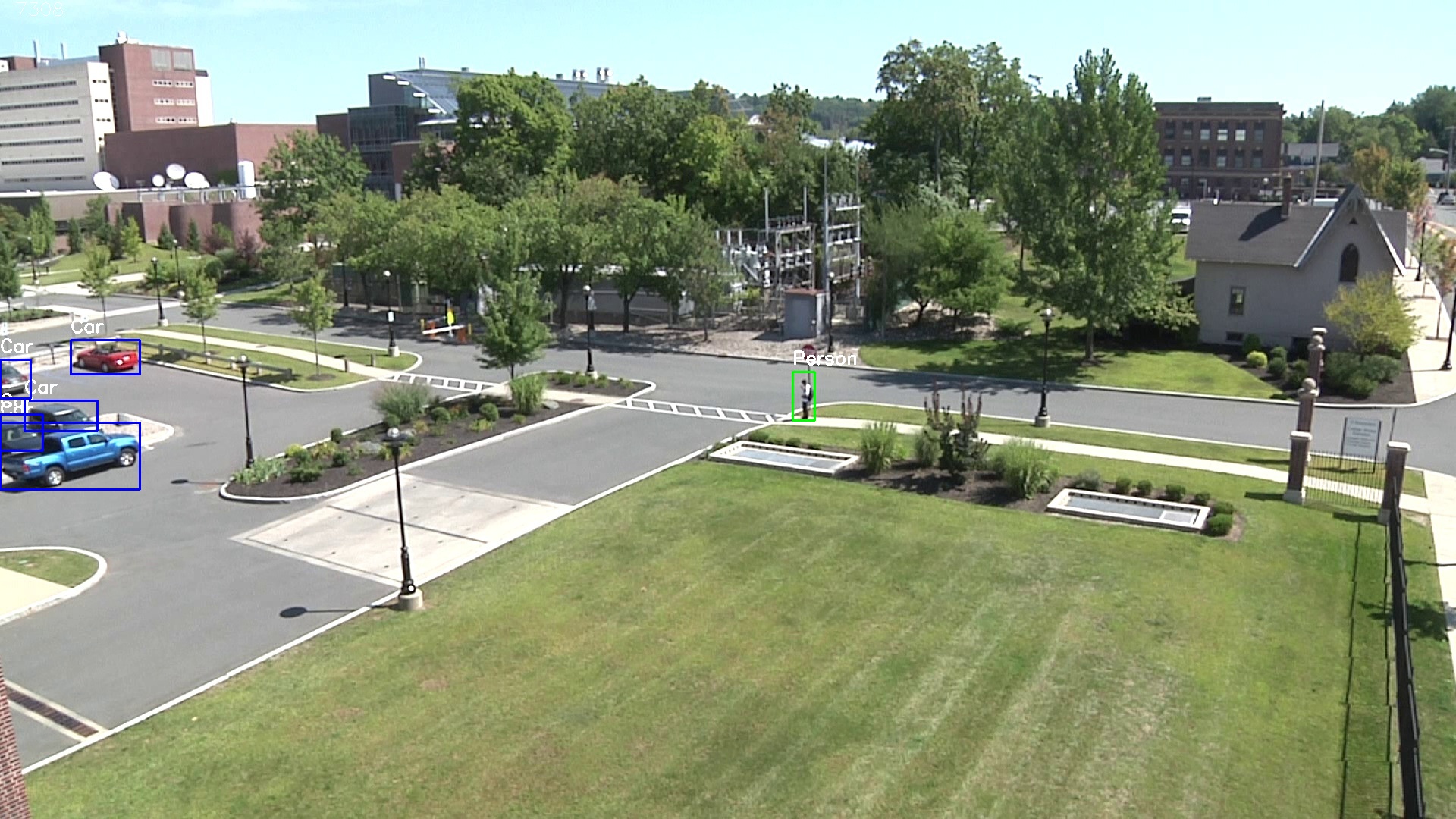}
	        \caption{Scene 7}
	        \label{fig:subfigC}
    \end{subfigure}
    \begin{subfigure}{0.23\linewidth}
	        \includegraphics[width=\linewidth]{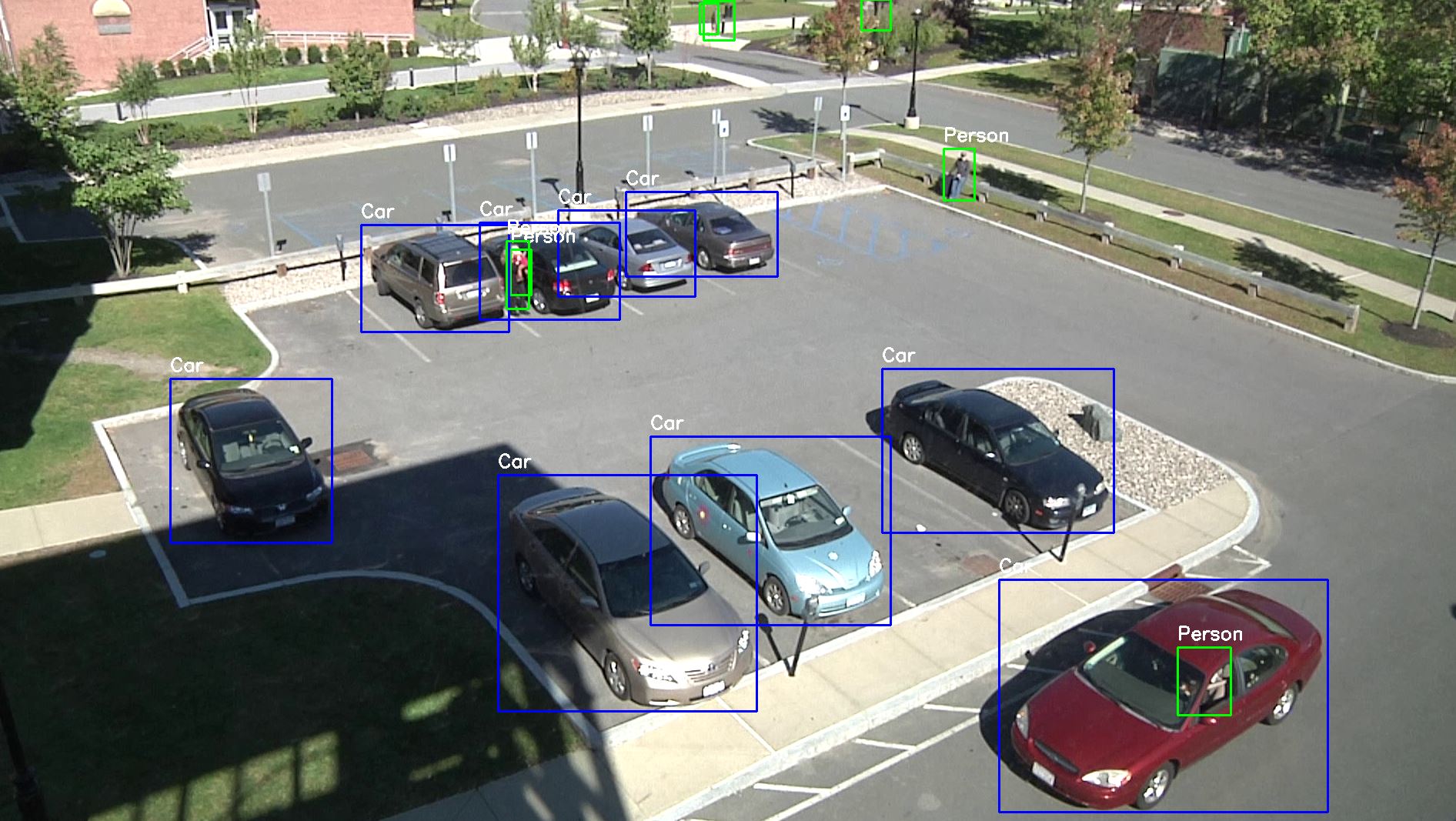}
	        \caption{Scene 8}
	        \label{fig:subfigC}
    \end{subfigure}
    \begin{subfigure}{0.23\linewidth}
	        \includegraphics[width=\linewidth]{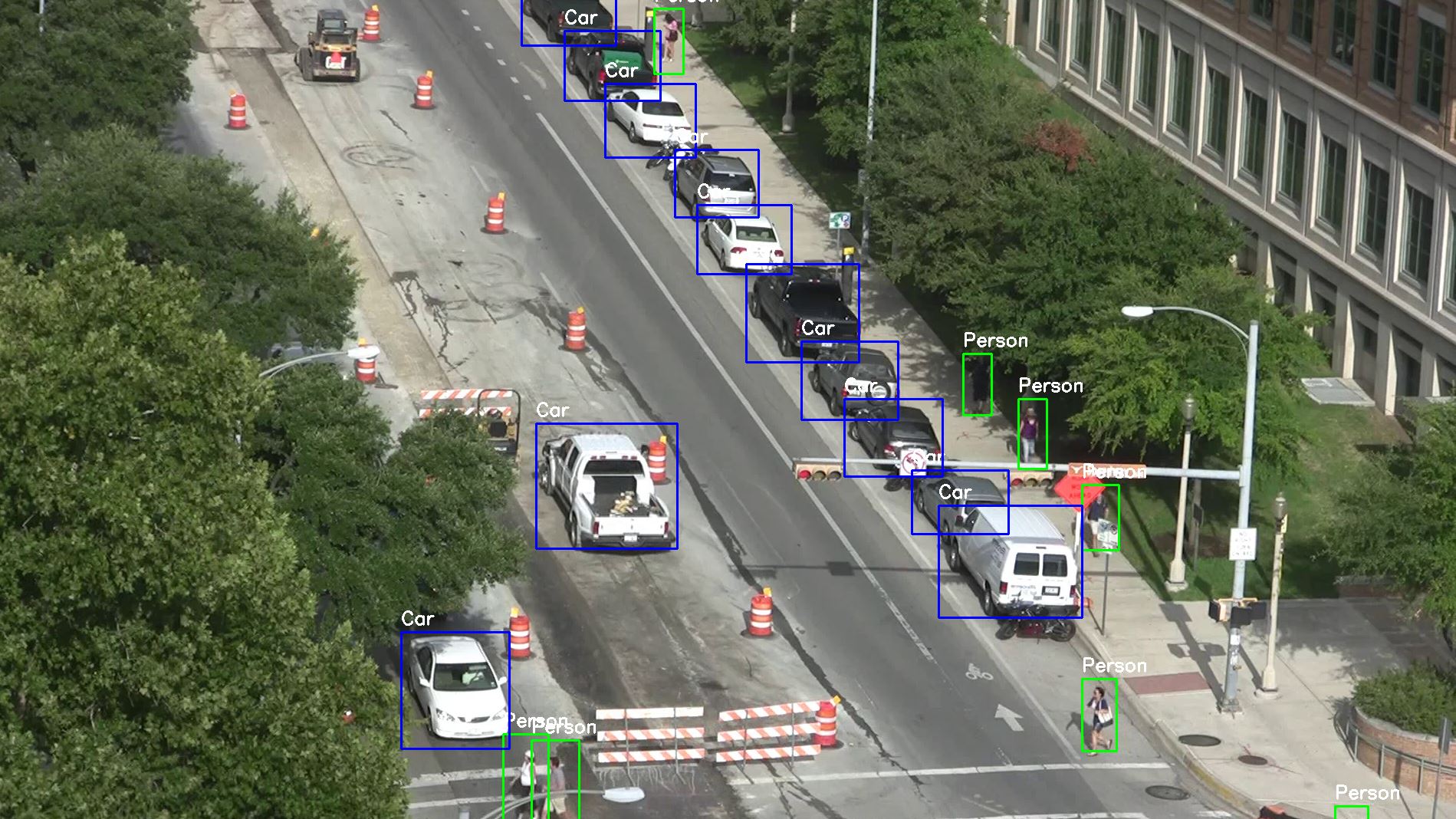}
	        \caption{Scene 9}
	        \label{fig:subfigC}
    \end{subfigure}
    \begin{subfigure}{0.23\linewidth}
	        \includegraphics[width=\linewidth]{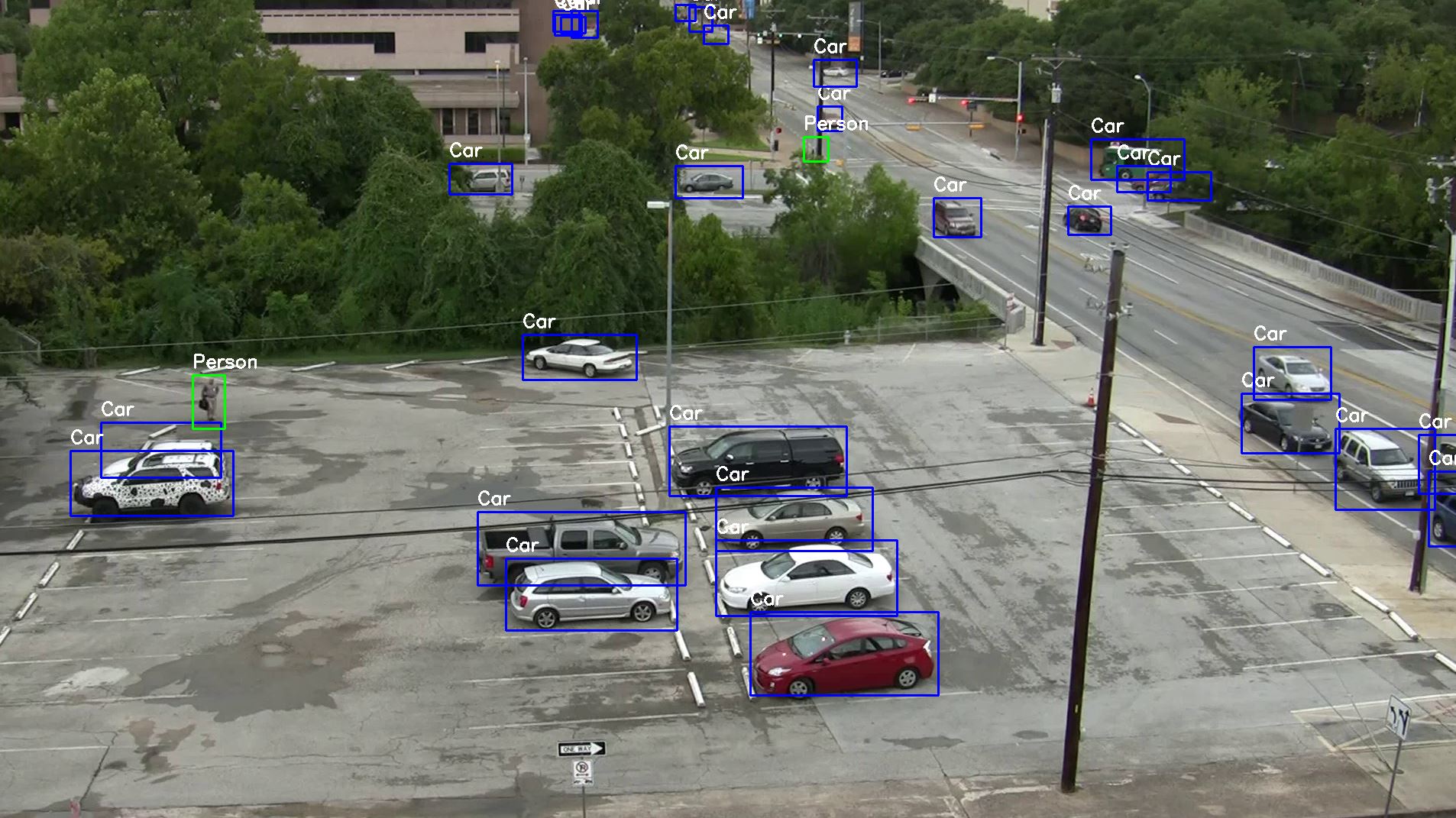}
	        \caption{Scene 10}
	        \label{fig:subfigC}
    \end{subfigure}
	\caption{Visual overview of different scenes from OD-VIRAT dataset and their annotations. Both OD-VIRAT Large and OD-VIRAT
Tiny dataset contain the above ten scenes.}
	\label{fig:scene_over}
\end{figure}

\begin{figure*}[t]
	\centering
	\includegraphics[width=\linewidth]{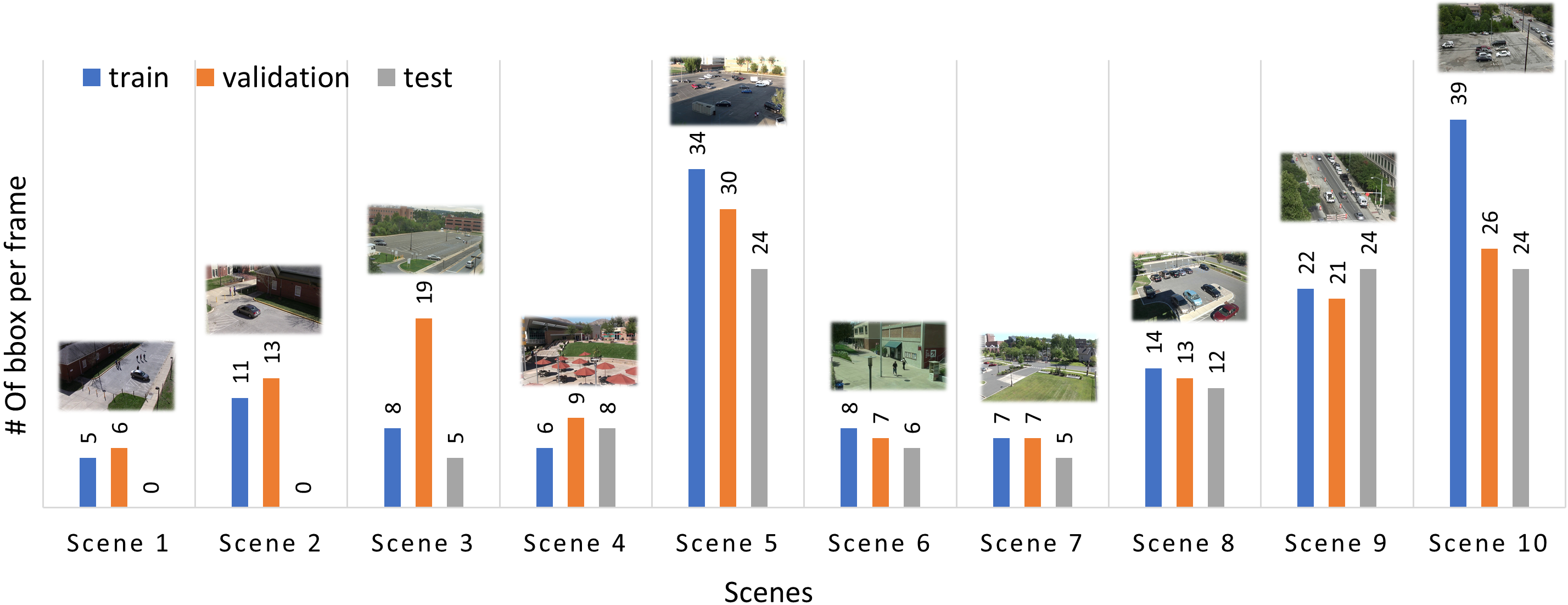}
	\caption{The number of bounding boxes per frame across 10 different scenes in train, validation, and test sets of OD-VIRAT Large and OD-VIRAT Tiny dataset.}
	\label{fig:bbox_per_frame}
\end{figure*}

Our data annotation process consists of three phases that include (1) bounding box coordinates retrieval from VIRAT Ground dataset annotations, (2) compilation of retrieved coordinates, (3) annotation conversion form {\fontfamily{qcr}\selectfont .txt} to coco \cite{lin2014microsoft}{\fontfamily{qcr}\selectfont .json} format. The first phase involves careful retrieval of frame-level bounding box coordinates of objects from the existing object-level annotation of VIRAT Ground dataset. The retrieved bounding box coordinates are then compiled in the second phase by applying the bounding boxes to their corresponding frames. Considering a scenario, where a single frame can have either no object or more than 1 object. Having this in mind, we first traversed bounding box coordinates using {\fontfamily{qcr}\selectfont frame\_id} and then applied them to their corresponding frames. After compilation, in third phase, the compiled bounding box coordinates are parsed from {\fontfamily{qcr}\selectfont .txt} format to coco {\fontfamily{qcr}\selectfont .json} format. Once converted, the prepared annotations for train, validation, and test sets are verified by applying them to corresponding frame data. The visual overview of each annotated scene from OD-VIRAT Large and OD-VIRAT Tiny datasets is depicted in Fig.~\ref{fig:scene_over}. Further, we computed the number of bounding boxes for objects per frame across each scene in train, validation, and test sets of OD-VIRAT Large and OD-VIRAT Tiny datasets, as shown in Fig.~\ref{fig:bbox_per_frame}.
\section{Methods For Benchmarking}
\label{sec:benchmarking_method}
This work seeks to investigate the impact of complex background and small-scale objects on the performance of state-of-the-art object detection architectures, their strength and limitations. To this end, two object detection benchmark datasets are created (i.e., the OD-VIRAT Large and OD-VIRAT Tiny described in the previous section), containing images with challenging properties i.e., complex background, occluded objects, and varying scale objects captured from significant height and distance. For benchmarking, we selected five state-of-the-art object detection architectures that include RTMDET \cite{lyu2022rtmdet}, YOLOX \cite{ge2021yolox}, RetinaNet \cite{lin2017focal}, DETR \cite{carion2020end}, and Deformable-DETR \cite{zhu2020deformable}. A brief overview and description of selected state-of-the-art object detection architectures are provided in the following subsections.

\subsection{RTMDET }
RTMDET \cite{lyu2022rtmdet},  presents an empirical investigation into real-time object detectors, exploring architectural choices and optimization techniques for high-speed performance. It examines the impact of components like backbone architectures, feature fusion, and inference strategies on efficiency and accuracy. Through extensive experimentation, authors provide insights into speed-accuracy trade-offs, offering practical guidance for designing efficient real-time detectors. RTMDET advances real-time object detection research, with implications for applications like autonomous driving, surveillance, and robotics. In this work, we investigate the performance of different RTMDET varients including RTMDET Tiny, RTMDET Small, RTMDET Medium, and RTMDET Large with CSPNeXt \cite{chen2024cspnext} backbone  on our OD-VIRAT Tiny dataset.

\subsection{YOLOX}
YOLOX \cite{ge2021yolox} is an innovative object detection architecture surpassing previous YOLO variants in speed and accuracy. It introduces CSPDarknet53 backbone for efficient feature extraction and FPN \cite{lin2017feature} for multi-scale fusion. YOLOX utilizes CSPNet \cite{wang2020cspnet} for better information flow and gradient propagation, achieving remarkable speed-accuracy trade-offs. Its decoupled head mechanism optimizes model performance, and anchor-free detection simplifies bounding box prediction without compromising accuracy. YOLOX's design choices contribute to state-of-the-art performance, incorporating training strategies like label smoothing and mixup augmentation for improved generalization and robustness. In this work, we examine the performance of different YOLOX varients including YOLOX Nano, YOLOX Tiny, YOLOX Small, YOLOX Medium, and YOLOX Large on our OD-VIRAT Tiny dataset.

\begin{table*}[t]
    \centering
    \caption{Configurations and experimental settings for each object detection model.}
    \resizebox{1\linewidth}{!}{
    \begin{tabular}{c|c|c|c|c|c}
    \toprule
         Configuration & RTMDET \cite{lyu2022rtmdet} & YOLOX \cite{ge2021yolox} & RetinaNet \cite{lin2017focal} & DETR \cite{carion2020end} & Deformable-DETR \cite{zhu2020deformable} \\
         \hline 
        optimizer & AdamW & SGD & SGD & AdamW & AdamW\\
        base learning rate  & 0.004 & 0.01 & 0.01 & 0.0001 & 0.0002   \\
        weight decay   & 0.05 & 0.0005 & 0.0001 & 0.0001 & 0.0001\\
        batch size   & 32\,$\mid$64$\mid$128 & 32\,$\mid$64$\mid$128 & 32\,$\mid$64$\mid1$28 & 32\,$\mid$ 64$\mid$128 & 32\,$\mid$64$\mid$128 \\
        optimizer momentum & \ding{55} & 0.9 & 0.9 & \ding{55} & \ding{55} \\
        parameters scheduler & CosineAnnealingLR & CosineAnnealingLR & MultiStepLR & MultiStepLR & MultiStepLR\\
        training epochs  & 50  & 50  & 50 & 50 & 50\\
    \bottomrule
    \end{tabular}}
    \label{tab:configs}
\end{table*}

\subsection{RetinaNet }
RetinaNet \cite{lin2017focal} a single-stage object detection architecture, integrates a Feature Pyramid Network (FPN) \cite{lin2017feature} with a backbone network and two subnetworks for classification and bounding box prediction. The backbone network extracts features from input images, creating feature maps of different resolutions, which are then combined by the FPN to form a multiscale feature pyramid. This facilitates the detection of objects of varying sizes. RetinaNet addresses challenges like unbalanced data and diverse object sizes through its specialized design, incorporating the Focal Loss function \cite{lin2017focal}. By leveraging these components, RetinaNet efficiently localizes objects in images, making it a powerful detector in object detection tasks. Our proposed benchmarking method includes the performance evaluation of RetinaNet architecture with three different backbones including ResNet18 \cite{he2016deep}, ResNet50 \cite{he2016deep}, and ResNet101 \cite{he2016deep} on our OD-VIRAT Tiny dataset.

\subsection{DETR}
Detection Transformer (DETR) \cite{carion2020end} revolutionized object detection by employing a transformer encoder-decoder architecture, eliminating anchor-based methods. It directly predicts object bounding boxes and classifications in a single pass, utilizing a set-based global objective function for end-to-end training without heuristic matching. With a multi-head self-attention mechanism, DETR effectively models long-range dependencies in images. Positional encodings are incorporated to provide spatial information to the attention mechanism. This transformer-based approach simplifies the detection pipeline while achieving impressive performance on various benchmarks, showcasing its transformative potential in object detection tasks. In this work, we evaluate the performance of DETR architecture with three different backbones including  ResNet18 \cite{he2016deep}, ResNet50 \cite{he2016deep}, and ResNet101 \cite{he2016deep} on our OD-VIRAT Tiny dataset.

\begin{figure*}[t]
    \centering
    \begin{subfigure}{0.32\textwidth}
        \includegraphics[width=\linewidth]{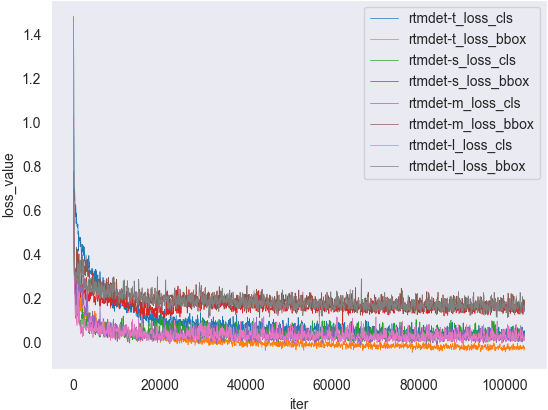}
        \caption{RTMDET }
        \label{fig:sub1}
    \end{subfigure}
    \begin{subfigure}{0.32\textwidth}
        \includegraphics[width=\linewidth]{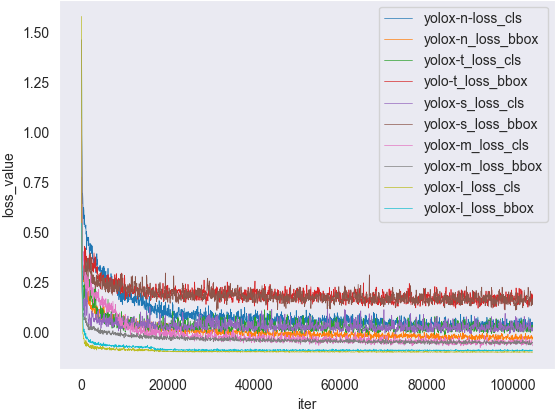}
        \caption{YOLOX }
        \label{fig:sub2}
    \end{subfigure}
    \begin{subfigure}{0.32\textwidth}
        \includegraphics[width=\linewidth]{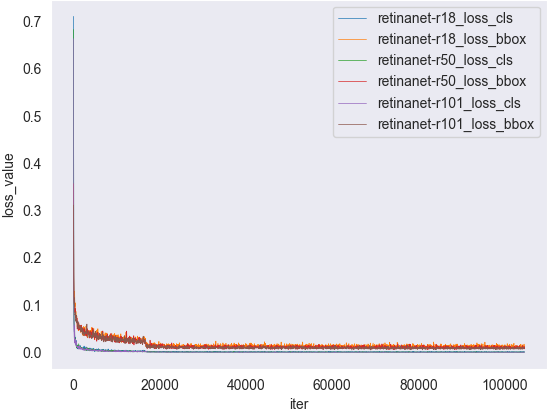}
        \caption{RetinaNet }
        \label{fig:sub3}
    \end{subfigure}
    \begin{subfigure}{0.32\textwidth}
        \includegraphics[width=\linewidth]{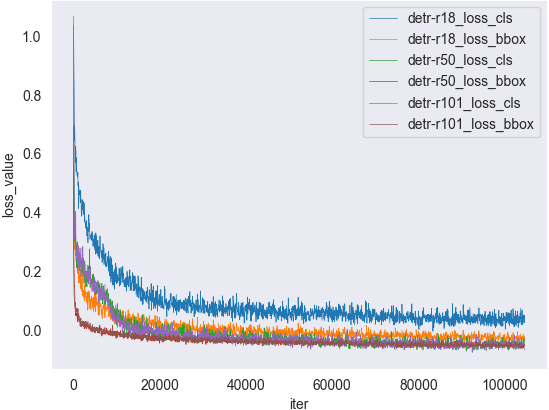}
        \caption{DETR }
        \label{fig:sub4}
    \end{subfigure}
    \begin{subfigure}{0.32\textwidth}
        \includegraphics[width=\linewidth]{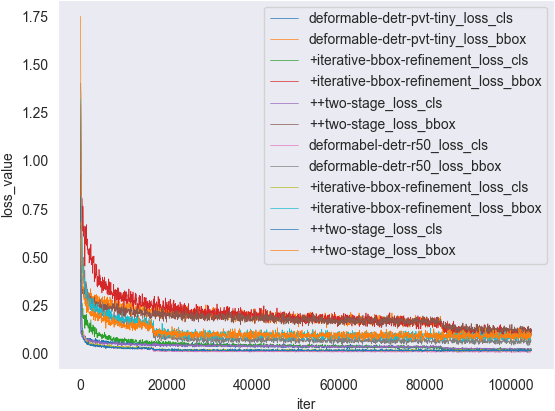}
        \caption{Deformable-DETR}
        \label{fig:sub5}
    \end{subfigure}
    \caption{Convergence curves (classification loss and bbox prediction loss) curves of RTMDET, YOLOX, RetinaNet, DETR, and Deformable-DETR model on OD-VIRAT Tiny dataset.}
    \label{fig:convergence}
\end{figure*}

\subsection{Deformable-DETR}
Deformable DETR \cite{zhu2020deformable}, an extension of DETR, integrates deformable attention mechanisms to better model object shapes and spatial relationships. It introduced deformable transformer layers that adjust attention sampling locations based on learned deformable offsets, improving object localization accuracy in the presence of deformations and occlusions. Maintaining end-to-end training, Deformable DETR enhances spatial information capture with deformable convolutional networks for feature extraction. This innovative design yields state-of-the-art performance on object detection benchmarks, showcasing its effectiveness in handling complex object scenarios. Deformable DETR holds promise for advancing object detection, particularly in scenarios with significant object deformations or occlusions.  Our proposed benchmarking method includes the performance evaluation of Deformable-DETR architecture and its variants with two different backbones that include PVT-Tiny \cite{wang2021pyramid} and ResNet50 \cite{he2016deep} on our OD-VIRAT Tiny dataset.

\section{Experimental Results}
\label{sec:experimetnal_results}
In this section, we present a detailed experimental evaluation of selected objected detection models on OD-VIRAT Tiny dataset. First, the implementation details and evaluation metrics used in our experiments are described. Next, we present the detailed quantitative and qualitative results of selected models OD-VIRAT Tiny dataset using different experimental settings. 
\subsection{Implementation Details}
\label{sec:implementation_details}
The experiments are performed on an Ubuntu-operated machine equipped with a 3.10GHz processor having 16 cores, 64GB of onboard RAM, and GeForce RTX 3090 8GB GPU operated with CUDA 12.1. Further, for implementation, PyTorch 2.1.0 is used with the backend of the MMDetection~\cite{mmdetection} codebase. It is worth mentioning here that, all experiments are conducted on the OD-VIRAT Tiny dataset due to the availability of limited resources and time. The detailed configurations and experimental settings for all selected objected detection models are listed in Table~\ref{tab:configs}. Further, the convergence curves of RTMDET, YOLOX, RetinaNet, DETR, and Deformable-DETR in terms of $loss_cls$ and $loss_bbox$ are depicted in Fig.~\ref{fig:convergence}.

\begin{table*}[t]
\caption{The obtained quantitative results of five state-of-the-art object detection models trained with and without pre-trained weights on OD-VIRAT tiny test set.}
	\label{Comparisons}
	\centering
	\resizebox{1\linewidth}{!}{
		\begin{tabular}{c|c|c|c|ccc|ccc|c|c}
			\toprule
			Method  & Citation & Pre-trained & Backbone &mAP &$mAP_{50}$ & $mAP_{75}$ & $mAP_{S}$  & $mAP_{M}$ & $mAP_{L}$ &\#Params &GFLOPs  \\

            \hline
            RTMDET Tiny &\multirow{4}{*}{\cite{lyu2022rtmdet}} & \ding{55} & CSPNeXt-P5 &16.5 &24.8 &19.3 &10.5 &30.4 &36.1 & 5M& 8 \\
            RTMDET Small & & \ding{55} & CSPNeXt-P5 &41.8 &58.1 &51.7 &15.9 &62.6 &73.0 &9M & 15\\
            RTMDET Medium & & \ding{55} & CSPNeXt-P5 &47.2 &61.6 &56.0 &18.6 &66.8 &76.5 & 25M& 39 \\
            RTMDET Large & &\ding{55} & CSPNeXt-P5 &50.1 &63.6 &60.2 &20.3 &69.1 &79.8 & 52M& 79 \\
			\hline
             RTMDET Tiny  &\multirow{4}{*}{\cite{lyu2022rtmdet}} &\ding{51} & CSPNeXt-P5 &27.1 &39.6 &29.3 &12.1 &40.6 &47.3 &5M &8  \\
            RTMDET Small & &\ding{51} & CSPNeXt-P5 &53.7 &70.3 &57.9 &18.9 &66.8 &88.1 &9M &15  \\
            RTMDET Medium & &\ding{51} & CSPNeXt-P5 &57.0 &72.4 &61.1 &28.1 &72.4 &90.7 &25M &39  \\
            RTMDET Large & &\ding{51} & CSPNeXt-P5 &59.2 &73.1 &63.4 &30.8 &73.7 &92.0 &52M &79 \\
            \hline
            YOLOX Nano  &\multirow{5}{*}{\cite{ge2021yolox}} &\ding{55} & CSPDarkNet &24.0 &36.6 &28.5 &16.0 &26.2 &61.2  &0.9M &1.2 \\
            YOLOX Tiny & &\ding{55} & CSPDarkNet &27.1 &42.3 &34.4 &19.0 &35.4 &65.9  &5M &7.5  \\
            YOLOX Small  & &\ding{55} & CSPDarkNet &36.3 &49.0 &39.4 &21.3 &41.1 &68.2  &8.9M &13.3  \\
            YOLOX Medium  & &\ding{55} & CSPDarkNet &42.1 &66.6 &48.1 &23.7 &50.9 &71.9  &25.2M &36.7  \\
            YOLOX Large  & &\ding{55} & CSPDarkNet &45.0 &72.6 &52.0 &28.2 &57.8 &74.5  &54.1M &77.6  \\
			\hline
            YOLOX Nano  &\multirow{5}{*}{\cite{ge2021yolox}} &\ding{51} & CSPDarkNet &31.2 &44.7 &38.2 &23.1 &33.3 &66.5  &0.9M &1.2  \\
            YOLOX Tiny  & &\ding{51} & CSPDarkNet &33.3 &47.2 &40.2 &24.7 &41.6 &70.4  &5M &7.5  \\
            YOLOX Small & &\ding{51} & CSPDarkNet &41.1 &52.1 &43.5 &27.2 &46.5 &73.3  &8.9M &13.3  \\
            YOLOX Medium & &\ding{51} & CSPDarkNet &47.3 &71.0 &52.3 &32.2 &57.6 &78.2 &25.2M &36.7 \\
            YOLOX Large  & &\ding{51} & CSPDarkNet &52.9 &78.7 &59.0 &36.1 &63.9 &80.0  &54.1M &77.6  \\
            \hline
            RetinaNet  &\multirow{3}{*}{\cite{lin2017focal}} &\ding{55} & ResNet18 &60.5 &77.2 &59.1 &44.4 &75.2 &83.1 &19M &96  \\
            RetinaNet  & &\ding{55} & ResNet50 &60.7 &78.6 &61.8 &43.8 &75.4 &84.1 &36M &128  \\
            RetinaNet  & &\ding{55} & ResNet101 &59.7 &77.6 &58.8 &44.9 &75.7 &82.6 &55M &176  \\
			\hline
            RetinaNet  &\multirow{3}{*}{\cite{lin2017focal}} &\ding{51} & ResNet18 &70.1 &84.0 &71.3 &58.5 &87.8 &96.5 &19M &96  \\
            RetinaNet  & &\ding{51} & ResNet50 &70.2 &84.2 &70.6 &60.3 &87.5 &96.1 &36M &128  \\
            RetinaNet  & &\ding{51} & ResNet101 &70.8 &85.5 &71.6 &59.4 &88.3 &96.7 & 55M &176  \\
			\hline
            DETR  &\multirow{3}{*}{\cite{carion2020end}} &\ding{55} & ResNet18 &32.7 &50.4 &39.2 &14.3 &48.8 &76.3 &28M &31  \\
            DETR  & &\ding{55} & ResNet50 &47.1 &53.8 &45.1 &23.9 &63.2 &82.4 &41M &60  \\
            DETR  & &\ding{55} & ResNet101 &42.3 &53.3 &44.6 &21.4 &57.7 &80.8 &60M &108  \\
            \hline
            DETR  &\multirow{3}{*}{\cite{carion2020end}} &\ding{51} & ResNet18 &47.4 &63.1 &49.4 &17.1 &56.6 &88.3 &28M &31 \\ 
            DETR  & &\ding{51} & ResNet50 &56.8 &72.4 &60.0 &28.3 &73.9 &91.8 &41M &60 \\
            DETR  & &\ding{51} & ResNet101 &53.3&73.6 &56.2 &26.5 &66.0 &90.8 &60M &108 \\
            \hline
            Deformable DETR  &\multirow{6}{*}{\cite{zhu2020deformable}}  &\ding{55} & PVT-Tiny &56.1 &64.4 &56.7 &44.3 &70.3 &81.2  & 24M &30.2  \\
			+ iterative bounding box refinement  & &\ding{55} & PVT-Tiny &59.4 &68.2 &60.2 &47.7 &74.4 &84.7 &24M &30.2\\
			
			++ two-stage Deformable DETR  & &\ding{55} & PVT-Tiny &62.7 &73.2 &62.4  &50.7 &77.2 &86.7 &24.9M &31.3 \\
            Deformable DETR  & &\ding{55} & ResNet50 &66.6 &76.1 &69.6 &51.3 &85.2 & 91.8 & 40M & 126\\
			+ iterative bounding box refinement  & &\ding{55} & ResNet50 &65.8 &77.3 &68.4 &50.9 &85.3 &92.1 & 40M & 126\\
			
			++ two-stage Deformable DETR  & &\ding{55} & ResNet50 &66.0 &78.8  &68.8  &50.5 &83.1 &91.9 & 40M & 126 \\
			\hline
            Deformable-DETR  &\multirow{6}{*}{\cite{zhu2020deformable}} &\ding{51} & PVT-Tiny &67.1 &81.7 &69.1&53.8&81.5& 91.6 &24M  & 30.2\\
			+ iterative bounding box refinement  & &\ding{51} & PVT-Tiny &67.3 &76.2 &68.0&52.6&83.4& 92.2 &24M  &30.2  \\
			++ two-stage-Deformable DETR  & &\ding{51} & PVT-Tiny & 68.3& 77.8 & 70.0 &54.9 &85.5 & 93.3 &24.9M  &31.3  \\
			Deformable-DETR  & &\ding{51} & ResNet50 &71.6 &90.7 &71.0&60.4&83.1& 96.5 & 40M &126 \\
			+ iterative bounding box refinement  & &\ding{51} & ResNet50 & \textit{73.8} & \textit{92.8} & \textit{76.1} & \textit{64.5} & \textit{87.6} & \textit{96.9} &40M &126 \\
			
			++ two-stage Deformable-DETR  & & \ding{51} & ResNet50 & \textbf{75.0}& \textbf{93.3} & \textbf{76.3} &\textbf{64.9} &\textbf{88.2} &\textbf{97.5} & 40M & 126 \\
        \bottomrule
			\end{tabular}}
\end{table*} 

\begin{table}[t]
\caption{Average $mPC$ values of RTMDET model (with \textbf{CSPNeXt-P5} backbone) under four image manipulation attacks, including Gaussian Noise, Motion Blur, Snow, and Elastic Transform with severity level 1 to 5.}
\centering
\scriptsize
\setlength{\tabcolsep}{2.5pt}
\begin{tabular}{c|ccc|ccc}
\toprule
 Method & $mAP$ & $mAP_{50}$ & $mAP_{75}$ & $mAP_{S}$ & $mAP_{M}$ & $mAP_{L}$ \\
\hline
RTMDET Tiny &  13.6 & 16.5 & 11.9 & 5.6 & 14.3 & 17.0  \\
RTMDET Small & 40.6 & 57.8 & 43.3 & 17.9 & 52.5 & 71.0 \\
RTMDET Medium & \textit{44.3}  & \textit{59.0} & \textit{49.4} & \textit{20.5} & \textit{56.4} & \textit{78.2}  \\
RTMDET Large & \textbf{47.0} &\textbf{ 60.4} & \textbf{52.7} & \textbf{23.6} & \textbf{58.9} & \textbf{79.1} \\
\bottomrule
\end{tabular}
\label{tab:mpc_rtmdet}
\end{table}%
\begin{table}[t]
\caption{Average $mPC$ values of YOLOX model (with \textbf{CSPDarkNet} backbone) under four image manipulation attacks, including Gaussian Noise, Motion Blur, Snow, and Elastic Transform with severity level 1 to 5.}
\centering
\scriptsize
\setlength{\tabcolsep}{2.5pt}
\begin{tabular}{c|ccc|ccc}
\toprule
Method & $mAP$ & $mAP_{50}$ & $mAP_{75}$ & $mAP_{S}$ & $mAP_{M}$ & $mAP_{L}$ \\
\hline
YOLOX Nano & 15.7 & 19.4 & 17.8 & 11.2 & 16.5 & 17.3  \\
YOLOX Tiny & 18.2 & 21.6 & 20.5 & 13.1 & 21.8 & 25.9 \\
YOLOX Small & 22.5  & 27.1 & 24.0 & 19.3 & 27.4 & 50.5  \\
YOLOX Medium & \textit{34.8} & \textit{48.5} & \textit{31.9} & \textit{24.4} & \textit{35.3} & \textit{54.1} \\
YOLOX Large & \textbf{41.1} & \textbf{55.6} & \textbf{36.8} & \textbf{28.9} & \textbf{42.0} & \textbf{57.9} \\
\bottomrule
\end{tabular}
\label{tab:mpc_yolox}
\end{table}%
\begin{table}[t]
\caption{Average $mPC$ values of RetinaNet model under four image manipulation attacks, including Gaussian Noise, Motion Blur, Snow, and Elastic Transform with severity level 1 to 5.}
\centering
\scriptsize
\setlength{\tabcolsep}{2.5pt}
\begin{tabular}{c|c|ccc|ccc}
\toprule
Method &  Backbone & $mAP$ & $mAP_{50}$ & $mAP_{75}$ & $mAP_{S}$ & $mAP_{M}$ & $mAP_{L}$ \\
\hline
RetinaNet & ResNet18 & 43.8 & 63.4 & 46.1 & 34.0 & 56.1 & 69.6  \\
RetinaNet & ResNet50 & \textit{44.5} & \textit{63.5} & \textit{46.7} & \textit{35.6} & \textit{56.3} & \textit{71.4} \\
RetinaNet & ResNet101 & \textbf{44.8}  & \textbf{64.2} & \textbf{47.2} & \textbf{36.5} & \textbf{57.0} & \textbf{71.9}  \\
\bottomrule
\end{tabular}
\label{tab:mpc_retinanet}
\end{table}%
\begin{table}[hbt!]
\caption{Average $mPC$ values of DETR model under four image manipulation attacks, including Gaussian Noise, Motion Blur, Snow, and Elastic Transform with severity level 1 to 5.}
\centering
\scriptsize
\setlength{\tabcolsep}{2.5pt}
\begin{tabular}{c|c|ccc|ccc}
\toprule
 Method &  Backbone & $mAP$ & $mAP_{50}$ & $mAP_{75}$ & $mAP_{S}$ & $mAP_{M}$ & $mAP_{L}$ \\
\hline
DETR & ResNet18 & 29.1 & 43.6 & 28.4 & 10.2 & 35.0 & 59.7  \\
DETR & ResNet50 & \textit{29.4} & \textit{50.0} & \textit{29.7} & \textit{14.7} & \textit{35.4} & \textit{61.5} \\
DETR & ResNet101 & \textbf{34.3}  & \textbf{54.9} & \textbf{33.7} & \textbf{16.7} & \textbf{44.3} & \textbf{63.3}  \\
\bottomrule
\end{tabular}
\label{tab:mpc_detr}
\end{table}
\subsection{Evaluation Metrics}
\label{sec:evaluation_metrics}
The quantitative performance of selected objected detection models is evaluated using Mean Average Precision (mAP) with different intersections over Union (IoU) thresholds and object scales. The mAP metrics with different IoU thresholds include $mAP$ (which represents $mAP_{50:95}$), $mAP_{50}$, and $mAP_{75}$, whereas the mAP metrics with different objects scales include $mAP_{S}$, $mAP_{M}$, and $mAP_{L}$.

\subsection{Quantitative Evaluation}
\label{sec:quantitative_evaluation}
This section provides a detailed quantitative evaluation of selected state-of-the-art object detection models, including RTMDET, YOLOX, RetinaNet, DETR, and Deformable-DETR on the OD-VIRAT Tiny dataset. Table~\ref{Comparisons} presents the quantitative performance of each state-of-the-art model (trained with and without pre-trained weights) in terms of six metrics, including $mAP$, $mAP_{50}$, $mAP_{75}$, $mAP_{S}$, $mAP_{M}$, and $mAP_{L}$ metric along with number of parameters and Giga Floating Point Operations ($GFLPOs$) values. As reported in Table~\ref{Comparisons}, Deformable-DETR with ResNet50 backbone obtains the best mAP values across each IoU and scale threshold. RetinaNet architecture with ResNet(18,50,101) backbones attains the second-best mAP values across each IoU threshold. It is worth noticing that, the RetinaNet with ResNet18 backbone shows remarkable performance by obtaining better $mAP_{75}$ and $mAP_{M}$ values than Deformable-DETR (with ResNet50 backbone) while having far less number of parameters and GFLOPs count. On the other hand, two-stage Deformable-DETR offers an average improvement of 3.1\% and 3.8\% over RetinaNet with ResNet50 and ResNet101, while having far less number of parameters and GFLOPs count. Further, the performance of RTMDET model is considerably low on OD-VIRAT Tiny test set in comparison with other models. While comparing YOLOX model with DETR, it can be notice that despite having far less number of parameters, YOLOX shows similar performance in comparison with DETR model. 

\begin{table*}[t]
\caption{Average $mPC$ values of Deformable-DETR model under four image manipulation attacks, including Gaussian Noise, Motion Blur, Snow, and Elastic Transform with severity level 1 to 5.}
\centering
\begin{tabular}{c|c|ccc|ccc}
\toprule
 Method &  Backbone & $mAP$ & $mAP_{50}$ & $mAP_{75}$ & $mAP_{S}$ & $mAP_{M}$ & $mAP_{L}$ \\
\hline
Deformable-DETR & PVT-Tiny & 35.3 & 50.5 & 37.5 & 22.7 & 45.0 & 58.1  \\
+iterative bounding box refinement & PVT-Tiny & 37.3 & 57.0 & 38.3 & 25.9 & 45.3 & 58.3 \\
++two-stage Deformable-DETR & PVT-Tiny & 40.6  & 58.1 & 42.6 & 29.0 & 51.5 & 63.1  \\
\hline
Deformable-DETR & ResNet50 & 38.6 & 55.2 & 39.9 & 26.3 & 46.6 & 66.3  \\
+iterative bounding box refinement & ResNet50 & 40.3 & 62.1 & 49.9 & 30.2 & 60.0 & 83.0 \\
++two-stage Deformable-DETR & ResNet50 & 49.0  & 62.8 & 50.8 & 30.7 & 61.3 & 87.0  \\
\bottomrule
\end{tabular}
\label{tab:mpc_defor_detr}
\end{table*}
\begin{table}[t]
\caption{The obtained quantitative results of RTMDET architecture (with \textbf{CSPNeXt-P5} backbone) in terms of $mAP$, $mAP_{50}$, $mAP_{75}$, $mAP_{S}$, $mAP_{M}$, and $mAP_{L}$ with three different batch sizes on OD-VIRAT Tiny dataset. BS represents the batch size.}
\centering
\scriptsize
\setlength{\tabcolsep}{2.5pt}
\begin{tabular}{c|c|ccc|ccc}
\toprule
Method & BS & $mAP$ & $mAP_{50}$ & $mAP_{75}$ & $mAP_{S}$ & $mAP_{M}$ & $mAP_{L}$ \\
\hline
RTMDET Tiny &\multirow{4}{*}{\rotatebox[origin=c]{0}{32}}& 15.3 & 24.3 & 18.6 & 9.4 & 29.8 & 35.1  \\
RTMDET Small &  & 41.6 & 57.7 & 51.0 & 15.3 & 62.3 & 72.6 \\
RTMDET Medium &  & 46.7 & 60.9 & 55.3 & 18.4 & 66.1 & 76.3  \\
RTMDET Large &  & 49.7 & 63.2 & 59.3 & 20.1 & 68.6 & 79.2 \\ 
\cline{1-1} \cline{2-8} 
RTMDET Tiny &  \multirow{4}{*}{\rotatebox[origin=c]{0}{64}}& 16.8 & 24.7 & 19.3 & 10.8 & 30.6 & 36.4  \\
RTMDET Small &  & 41.9 & 58.3 & 51.8 & 15.8 & 62.5 & 73.1 \\
RTMDET Medium &  & 47.2 & 61.6 & 56.2 & 18.7 & 66.8 & 76.4  \\
RTMDET Large &  &\textit{50.2} & \textit{63.7} & \textit{59.8} & \textit{20.4} & \textit{69.1} & \textit{79.8} \\ 
\cline{1-1} \cline{2-8} 
RTMDET Tiny & \multirow{4}{*}{\rotatebox[origin=c]{0}{128}}& 17.40 & 25.4 & 20.2 & 11.5 & 31.0 & 36.8  \\
RTMDET Small &  & 42.0 & 58.5 & 52.4 & 16.6 & 63.2 & 73.5 \\
RTMDET Medium &  & 47.9 & 62.3 & 56.7 & 18.9 & 67.5 & 77.0 \\
RTMDET Large &  &\textbf{50.6} & \textbf{64.0} & \textbf{61.5} & \textbf{20.6} & \textbf{69.6} &\textbf{80.4} \\
\bottomrule
\end{tabular}
\label{tab:bs_rtmdet}
\end{table}
\begin{table}[hbt!]
\caption{The obtained quantitative results of YOLOX architecture (with \textbf{CSPDarkNet} backbone) in terms of $mAP$, $mAP_{50}$, $mAP_{75}$, $mAP_{S}$, $mAP_{M}$, and $mAP_{L}$ with three different batch sizes on OD-VIRAT Tiny dataset.  BS represents the batch size.}
\centering
\scriptsize
\setlength{\tabcolsep}{2.5pt}
\begin{tabular}{c|c|ccc|ccc}
\toprule
Method & BS & $mAP$ & $mAP_{50}$ & $mAP_{75}$ & $mAP_{S}$ & $mAP_{M}$ & $mAP_{L}$ \\
\hline
YOLOX Nano & \multirow{5}{*}{\rotatebox[origin=c]{0}{32}}& 24.0 & 38.7 & 30.2 & 16.7 & 26.9 & 62.3  \\
YOLOX Tiny &  & 26.7 & 41.8 & 33.4 & 18.6 & 34.8 & 65.6 \\
YOLOX Small &  & 35.9 & 48.5 & 39.5 & 21.4 & 40.2 & 68.1  \\
YOLOX Medium &  & 42.8 & 67.0 & 48.1 & 23.4 & 50.3 & 70.4  \\ 
YOLOX Large &  &\textit{44.1} & \textit{72.8} & \textit{50.3} & \textit{26.7} & \textit{57.2} & \textit{74.7} \\ 
\cline{1-1} \cline{2-8}
YOLOX Nano &  \multirow{5}{*}{\rotatebox[origin=c]{0}{64}}& 23.6 & 36.2 & 29.4 & 16.0 & 27.3 & 61.5  \\
YOLOX Tiny &  & 27.0 & 42.6 & 34.8 & 19.1 & 35.6 & 66.0 \\
YOLOX Small &  & 36.3 & 49.1 & 39.0 & 21.1 & 41.5 & 68.0  \\
YOLOX Medium &  & 42.1 & 66.5 & 48.4 & 24.0 & 51.7 & 70.8 \\ 
YOLOX Large &  &\textbf{47.6} & \textbf{73.2} &\textbf{56.3} & \textbf{32.3} & \textbf{61.9} & \textbf{76.8} \\ 
\cline{1-1} \cline{2-8}
YOLOX Nano &  \multirow{5}{*}{\rotatebox[origin=c]{0}{128}}& 24.5 & 34.9 & 26.0 & 15.3 & 27.4 & 59.9 \\
YOLOX Tiny &  & 27.8 & 42.7 & 35.1 & 19.5 & 36.0 & 66.3 \\
YOLOX Small &  & 36.8 & 49.4 & 39.7 & 21.6 & 41.8 & 68.5 \\
YOLOX Medium &  & 41.6 & 66.3 & 48.0 & 23.9 & 50.8 & 70.2 \\
YOLOX Large &  & 43.3 & 72.0 & 49.6 & 25.8 & 54.5 & 72.1 \\ 
\bottomrule
\end{tabular}
\label{tab:bs_yolox}
\end{table}
\begin{table}[hbt!]
\caption{The obtained quantitative results of RetinaNet architecture in terms of $mAP$, $mAP_{50}$, $mAP_{75}$, $mAP_{S}$, $mAP_{M}$, and $mAP_{L}$ with three different backbones and batch sizes on OD-VIRAT Tiny dataset. BS represents the batch size.}
\centering
\setlength{\tabcolsep}{2.5pt}
\begin{tabular}{c|c|ccc|ccc}
\toprule
Method &  BS & $mAP$ & $mAP_{50}$ & $mAP_{75}$ & $mAP_{S}$ & $mAP_{M}$ & $mAP_{L}$ \\
\hline
RetinaNet-r18 &\multirow{3}{*}{\rotatebox[origin=c]{0}{32}} & 59.5 & 75.7 & 58.4 & 44.0 & \textit{76.3} & 83.1  \\
RetinaNet-r50 & & 59.4 & 78.2 & \textit{61.6} & 44.7 & 75.0 & \textit{84.3} \\
RetinaNet-r101 & & 59.2 & 77.5 & 58.8 & 43.2 & 75.6 & 82.0  \\ \hline
RetinaNet-r18 & \multirow{3}{*}{\rotatebox[origin=c]{0}{64}}& 60.8 & 77.2 & 59.0 & \textit{45.5} & 75.1 & 83.8  \\
RetinaNet-r50 & & \textbf{61.5} & \textit{78.7} & \textbf{62.7} & 43.5 & 75.4 & 83.7 \\
RetinaNet-r101 & & 59.7 & 78.1 & 58.5 & 45.4 & 75.2 & 82.4  \\ \hline
RetinaNet-r18 & \multirow{3}{*}{\rotatebox[origin=c]{0}{128}}& \textit{61.3} & 78.5 & 60.1 & 43.7 & 74.3 & 82.6  \\
RetinaNet-r50 & & 61.2 & \textbf{79.0} & 61.3 & 43.2 & 75.9 & \textbf{84.5} \\
RetinaNet-r101 & &  60.4 & 77.4 & 59.3 & \textbf{46.2} & \textbf{76.5} & 83.5 \\ 
\bottomrule
\end{tabular}
\label{tab:bs_retinanet}
\indent{\footnotesize{\textsuperscript{\mbox{*}}  r-18, r50, and r-101 represent ResNet18, ResNet50, and ResNet101 backbones, repsectively.}}
\end{table}
\begin{table}[hbt!]
\caption{The obtained quantitative results of DETR architecture in terms of $mAP$, $mAP_{50}$, $mAP_{75}$, $mAP_{S}$, $mAP_{M}$, and $mAP_{L}$ with three different backbones and batch sizes on OD-VIRAT Tiny dataset.}
\centering
\setlength{\tabcolsep}{2.5pt}
\begin{tabular}{c|c|ccc|ccc}
\toprule
Method &  BS & $mAP$ & $mAP_{50}$ & $mAP_{75}$ & $mAP_{S}$ & $mAP_{M}$ & $mAP_{L}$ \\
\hline
DETR-r18 & \multirow{3}{*}{\rotatebox[origin=c]{0}{32}} & 33.1 & 51.4 & 38.0 & 19.5 & 49.7 & 77.2  \\
DETR-r50 & & \textbf{47.8} & \textbf{54.7} & \textit{45.9} & 23.6 & \textit{64.0} & \textit{82.5} \\
DETR-r101 & & 41.7 & 53.1 & 43.9 & \textit{24.5} & 55.4 & 81.0  \\ \hline
DETR-r18 & \multirow{3}{*}{\rotatebox[origin=c]{0}{64}}& 32.6 & 50.9 & 38.4 & 11.7 & 48.3 & 75.8  \\
DETR-r50 & & \textit{47.2} & 53.8 & \textbf{46.3} & 21.5 & \textbf{64.5} & \textbf{83.0} \\
DETR-r101 & & 43.2 & \textit{54.6} & 44.7 & 15.7 & 59.6 & 81.6  \\ \hline
DETR-r18 & \multirow{3}{*}{\rotatebox[origin=c]{0}{128}}& 32.4 & 49.0 & 37.6 & 11.9 & 48.4 & 76.1  \\
DETR-r50 & & 46.5 & 53.1 & 43.1 & \textbf{26.8} & 61.3 & 81.8 \\
DETR-r101 & &  42.0 & 52.3 & 45.4 & 24.0 & 58.2 & 79.8 \\ 
\bottomrule
\end{tabular}
\label{tab:bs_detr}\\
\indent{\footnotesize{\textsuperscript{\mbox{*}}  r-18, r50, and r-101 represent ResNet18, ResNet50, and ResNet101 backbones, repsectively.}}
\end{table}
\begin{table*}[t]
\caption{The obtained quantitative results of Deformable DETR architecture in terms of $mAP$, $mAP_{50}$, $mAP_{75}$, $mAP_{S}$, $mAP_{M}$, and $mAP_{L}$ with two different backbones and three different batch sizes on OD-VIRAT Tiny dataset.}
\centering
\begin{tabular}{c|c|c|ccc|ccc}
\toprule
Method &  Backbone & Batch Size & $mAP$ & $mAP_{50}$ & $mAP_{75}$ & $mAP_{S}$ & $mAP_{M}$ & $mAP_{L}$ \\
\hline
Deformable-DETR & \multirow{9}{*}{\rotatebox[origin=c]{90}{PVT Tiny}} &\multirow{3}{*}{\rotatebox[origin=c]{0}{32}} & 55.8 & 63.7 & 56.3 & 43.8 & 69.8 & 81.0 \\
+ iterative bounding box refinement & && 59.4 & 68.3 & 60.4 & 47.4 & 74.1 & 84.5\\
++ two-stage Deformable-DETR & && 62.3 & 72.9 & 62.3 & 50.6 & 76.8 & 86.3  \\ 
\cline{1-1} \cline{3-9}
Deformable-DETR &  &\multirow{3}{*}{\rotatebox[origin=c]{0}{64}}& 56.0 & 64.6 & 57.0 & 44.3 & 70.5 & 81.6  \\
+ iterative bounding box refinement &  && 59.0 & 67.9 & 59.6 & 48.0 & 74.8 & 84.9 \\
++ two-stage Deformable-DETR &  && 62.9 & 73.5 & 62.9 & 51.1 & 77.3 & 86.8  \\
\cline{1-1} \cline{3-9}
Deformable-DETR &  &\multirow{3}{*}{\rotatebox[origin=c]{0}{128}}& 56.5 & 64.1 & 56.8 & 44.9 & 70.7 & 81.1  \\
+ iterative bounding box refinement &  && 59.8 & 68.6 & 60.7 & 47.8 & 74.3 & 84.7 \\
++ two-stage Deformable-DETR &  && 63.0 & 73.3 & 63.0 & 50.4 & 77.6 & 87.2 \\ \hline
Deformable-DETR & \multirow{9}{*}{\rotatebox[origin=c]{90}{ResNet50}} &\multirow{3}{*}{\rotatebox[origin=c]{0}{32}} & 66.1 & 75.4 & 69.3 & 49.9 & 85.0 & 91.3  \\
+ iterative bounding box refinement & && \textit{66.7} & 77.5 & 69.6 & 52.0 & \textbf{86.3} & 92.1 \\
++ two-stage Deformable-DETR & && 66.5 & \textit{79.1} & 69.1 & 49.0 & 83.6 & 91.6 \\ 
\cline{1-1} \cline{3-9}
Deformable-DETR &  &\multirow{3}{*}{\rotatebox[origin=c]{0}{64}}& 66.4 & 75.7 & \textbf{69.7} & 49.6 & \textit{85.5} & \textit{92.6}  \\
+ iterative bounding box refinement &  && 65.9 & 78.2 & 68.2 & 48.6 & 85.4 & 92.4 \\
++ two-stage Deformable-DETR &  && 66.2 & \textbf{80.7} & 68.3 & 49.7 & 83.4 & \textbf{92.8}  \\
\cline{1-1} \cline{3-9}
Deformable-DETR &  &\multirow{3}{*}{\rotatebox[origin=c]{0}{128}}& \textbf{67.3} & 77.2 & \textbf{70.3} & \textbf{55.1} & 85.3 & 91.7  \\
+ iterative bounding box refinement &  && 65.0 & 76.2 & 67.6 & 52.3 & 84.4 & 91.8  \\
++ two-stage Deformable-DETR &  && 65.4 & 76.6 & 68.8 & \textit{53.0} & 82.5 & 91.3 \\
\bottomrule
\end{tabular}
\label{tab:bs_def_detr}
\end{table*}
\subsection{Robustness Analysis}
\label{sec:robustness_analysis}
To study the effect of different image manipulation (perturbation) schemes on the performance of selected object detection models, we evaluate the performance of each model on test set of OD-VIRAT Tiny perturbed with four different types of noise, including $Gaussian\;Noise$, $Motion\; Blur$, $Snow$, and $Elastic\;Transform$ with severity level 1 to 5. It is worth mentioning here that, in model robustness evaluation experiments we tested only those models that were trained with pre-trained weights. The best mAP values are emphasized in (bold) and the second-best mAP values are in (italic). The obtained Mean Performance under Corruption (mPC) results of each model are presented in Table~\ref{tab:mpc_rtmdet}-\ref{tab:mpc_defor_detr}. The mPC values given in Table~\ref{tab:mpc_rtmdet} shows the performance reduction of RTMDET architecture across each metric when testing on perturbed images. For instance, the average performance ($mAP:mAP_{L}$) reduction of RTMDET Tiny, Small, Medium, and Large on perturbed images is 19.4\%, 12.1\%, 12.3\%, and 11.7, respectively. Similarly, the average performance ($mAP:mAP_{L}$) reduction of YOLOX Nano, Tiny, Small, Medium, and Large on perturbed images is 23.1\%, 22.7\%, 18.8\%, 18.2\%, and 18.0, respectively, as shown in Table~\ref{tab:mpc_yolox}.\\
\indent The similar trend can be found in Table~\ref{tab:mpc_retinanet}, exhibiting the average performance ($mAP:mAP_{L}$) reduction of RetinaNet-ResNet18, RetinaNet-ResNet50, and RetinaNet-ResNet101 by 25.8\%, 25.1\%, and 25.1\%, respectively. While testing DETR architecture on a perturbed images, its average performance ($mAP:mAP_{L}$) is reduced by 19.3\%, 27\%, and 19.8\% for DETR-ResNet18, DETR-ResNet50, and DETR-ResNet101, respectively, as shown in Table~\ref{tab:mpc_detr}. Finally, the results reported in Table~\ref{tab:mpc_defor_detr}, exhibiting the average performance ($mAP:mAP_{L}$) reduction of Deformable-DETR with PVT-Tiny and ResNet backbone by 29.7\% and 28.9, respectively.

\begin{table}[t]
\caption{The obtained FPS values of different variants of RTMDET architecture.}
\centering
\begin{tabular}{c|c|c|c}
\toprule
Method & Backbone & FPS & Avg FPS \\
\hline
RTMDET Tiny &\multirow{4}{*}{\rotatebox[origin=c]{0}{CSPNeXt-P5}}& 86.5 & \multirow{4}{*}{\rotatebox[origin=c]{0}{81.6}}  \\
RTMDET Small &  & 84.3 &  \\
RTMDET Medium &  & 82.7 &    \\
RTMDET Large &  & 73.0 &   \\ 
\bottomrule
\end{tabular}
\label{tab:fps_rtmdet}
\end{table}
\begin{table}[t]
\caption{The obtained FPS values of different variants of YOLOX architecture.}
\centering
\begin{tabular}{c|c|c|c}
\toprule
Method & Backbone & FPS & Avg FPS \\
\hline
YOLOX Nano &\multirow{4}{*}{\rotatebox[origin=c]{0}{CSPNeXt-P5}}& 281.0 & \multirow{4}{*}{\rotatebox[origin=c]{0}{256.6}}  \\
YOLOX Tiny &  & 266.3 &  \\
YOLOX Small &  & 248.4 &    \\
YOLOX Medium &  & 247.4 & \\
YOLOX Large &  & 240.2 & \\ 
\bottomrule
\end{tabular}
\label{tab:fps_yolox}
\end{table}
\begin{table}[t]
\caption{The obtained FPS values of different variants of RetinaNet architecture.}
\centering
\begin{tabular}{c|c|c|c}
\toprule
Method & Backbone & FPS & Avg FPS \\
\hline
RetinaNet & ResNet18 & 152.6 & \multirow{4}{*}{\rotatebox[origin=c]{0}{149.8}}  \\
RetinaNet & ResNet50 & 148.8 &  \\
RetinaNet & ResNet101 & 148.1 &    \\
\bottomrule
\end{tabular}
\label{tab:fps_retinanet}
\end{table}
\begin{table}[hbt!]
\caption{The obtained FPS values of different variants of DETR architecture.}
\centering
\begin{tabular}{c|c|c|c}
\toprule
Method & Backbone & FPS & Avg FPS \\
\hline
DETR & ResNet18 & 151.7 & \multirow{4}{*}{\rotatebox[origin=c]{0}{148.4}}  \\
DETR & ResNet50 & 146.9 &  \\
DETR & ResNet101 & 146.7 &    \\
\bottomrule
\end{tabular}
\label{tab:fps_detr}
\end{table}
\begin{table}[hbt!]
\caption{The obtained FPS values of different variants of Deformable-DETR architecture.}
\centering
\begin{tabular}{c|c|c|c}
\toprule
Method & Backbone & FPS & Avg FPS \\
\hline
Deformable-DETR & \multirow{3}{*}{\rotatebox[origin=c]{0}{PVT Tiny}} & 148.2 & \multirow{4}{*}{\rotatebox[origin=c]{0}{145.3}}  \\
+ iterative bounding box refinement &  & 144.3 &  \\
++ two-stage Deformable-DETR &  & 143.4 &    \\
\midrule
Deformable-DETR & \multirow{3}{*}{\rotatebox[origin=c]{0}{ResNet50}} & 148.5 & \multirow{4}{*}{\rotatebox[origin=c]{0}{145.6}}  \\
+ iterative bounding box refinement &  & 145.2 &  \\
++ two-stage Deformable-DETR &  & 143.2 &    \\
\bottomrule
\end{tabular}
\label{tab:fps_defor_detr}
\end{table}

\subsection{Effect of Different Batch Sizes}
\label{sec:batch_size_effect}
This section presents the quantitative results of selected object detection models trained with different batch sizes (including 32, 64, and 128) on OD-VIRAT Tiny dataset. The obtained quantitative results with different batch sizes are presented in Table~\ref{tab:bs_rtmdet}-\ref{tab:bs_def_detr}. It is worth mentioning here that, these experiments are conducted using models trained without the utilization of pre-training weights. The best mAP values are emphasized in (bold) and the second-best mAP values are in (italic). Table~\ref{tab:bs_rtmdet} presents the RTMDET performance with different batch sizes on OD-VIRAT Tiny dataset. As can be seen in Table~\ref{tab:bs_rtmdet} RTMDET Large obtains the best mAP values with a batch size of 128 and the second-best mAP values with a batch size of 64. YOLOX obtains the best mAP values with a batch size of 64 and the second-best mAP values with a batch size of 32, as shown in Table~\ref{tab:bs_yolox}. The RetinaNet performance with three different backbones and batch sizes are presented in Table~\ref{tab:bs_retinanet}, demonstrating the best performance ($mAP$, $mAP_{50}$, $mAP_{75}$, and $mAP_{L}$) of RetinaNet architecture (using ResNet50 backbone) with batch size 64 and 128. As shown in Table~\ref{tab:bs_detr}, DETR architecture (using ResNet50 backbone) obtains the best mAP values with batch size 32 ($mAP$ and $mAP_{50}$) and batch size 64 ($mAP_{75}$, $mAP_{M}$, and $mAP_{L}$). Finally, Deformable-DETR architecture (using ResNet50 backbone) has the best mAP values with a batch size of 64 ($mAP$, $mAP_{75}$, and $mAP_{S}$) and a batch size 128 ($mAP_{50}$ and $mAP_{L}$), as shown in Table~\ref{tab:bs_def_detr}.

\begin{figure*}[t]
	\centering
	\includegraphics[width=\linewidth]{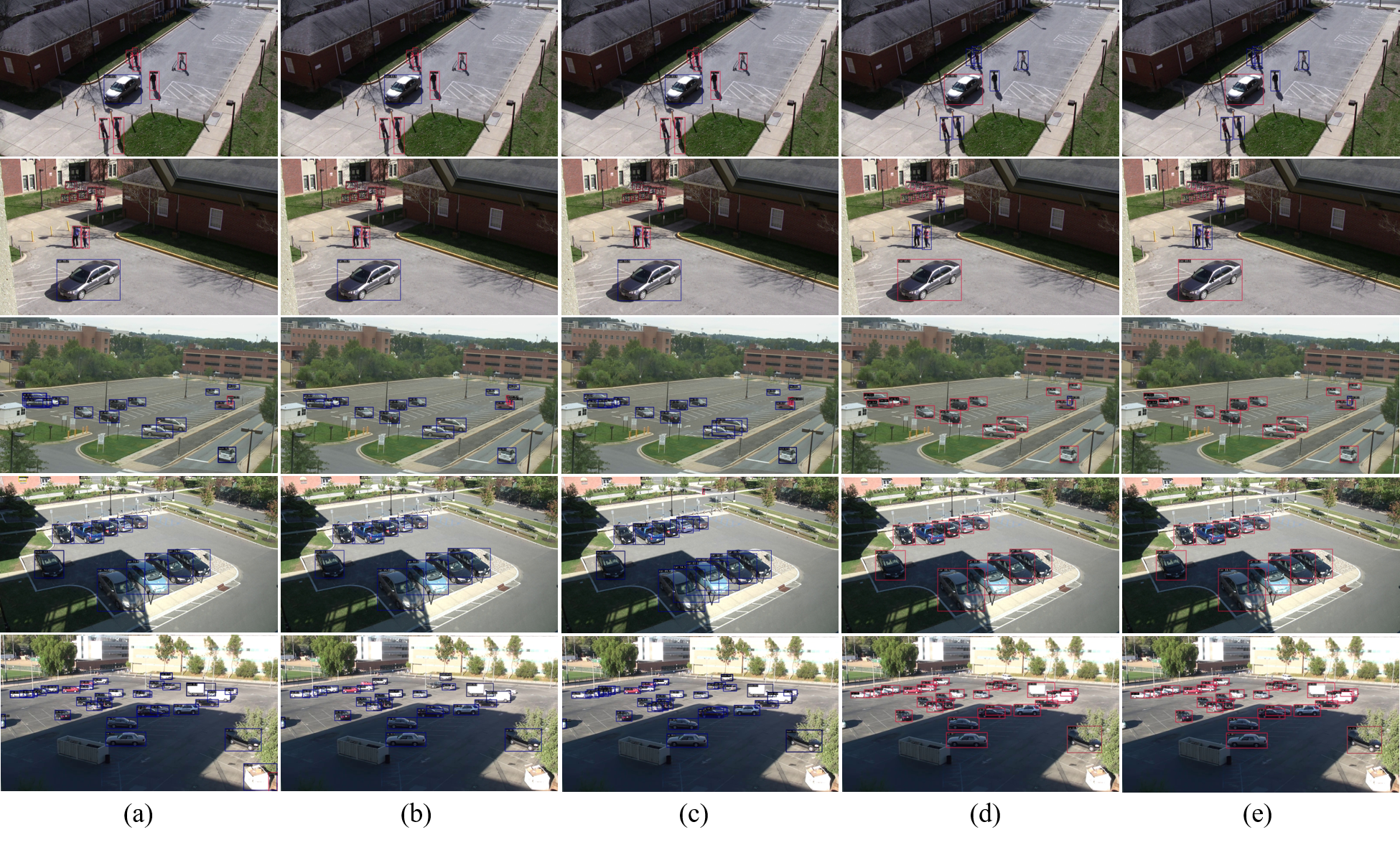}
	\caption{Visual comparative analysis of selected object detection models on five test images. (a) RTMDET, (b) YOLOX, (c) RetinaNet, (d) DETR, and (e) Deformable-DETR.}
	\label{fig:visual_results}
\end{figure*}
\subsection{FPS Benchmarking}
\label{sec:fps_benchmarking}
To evaluate the real-time detection efficiency, we computed the FPS of each model included in our experimental settings on the test set of OD-VIRAT Tiny dataset. The obtained FPS bechmarking results of RTMDET, YOLOX, RetinaNet, DETR, and Deformable-DETR are presented in Table~\ref{tab:fps_rtmdet}, \ref{tab:fps_yolox}, \ref{tab:fps_retinanet}, \ref{tab:fps_detr}, and \ref{tab:fps_defor_detr}, respectively. The tabulated results demonstrate the real-time efficiency of YOLOX architecture thereby obtaining the best average FPS of 256.6, followed by RetinaNet architecture having the second-best FPS of 149.8. Approximately similar FPS is obtained by DETR architecture, having FPS of 148.4. The Deformable-DETR architecture obtains the FPS of 145.3 with PVT Tiny backbone and FPS of 145.6 with ResNet50 backbone. The RTMDET architecture has the lowest FPS amongst selected object detection models, obtaining the FPS of 81.6.
\subsection{Qualitative Evaluation}
\label{sec:qualitative_evaluation}
It is also essential to visually investigate the performance of selected state-of-the-art object detection models on OD-VIRAT Tiny dataset. For qualitative evaluation, we tested each model on a set of five images from the test set of OD-VIRAT Tiny dataset. Each image represents a different scene, where each scene varies in terms of number of objects, distance of objects from a camera, angle of view, and degree of occlusion between objects. The obtained visual object detection results of each model on five different images are depicted in Fig.~\ref{fig:visual_results}. From the visual results, it can be perceived that the detection output of RetinaNet contains some redundant prediction of bounding boxes for images in first, third and fourth row. Similarly, the detection output of YOLOX and RTMDET models has redundant prediction of bounding boxes in first and third row. It is evident from the visual results in fourth and fifth column, that the detection output of DETR and Deformable-DETR are flawless, providing precise prediction of bounding boxes. Thus, showcasing the effectiveness of transformer-based architectures and their emerging characteristics for complex object detection task in realistic surveillance scene imagery.\\
\begin{figure*}[hbt!]
	\centering
	\includegraphics[width=\linewidth]{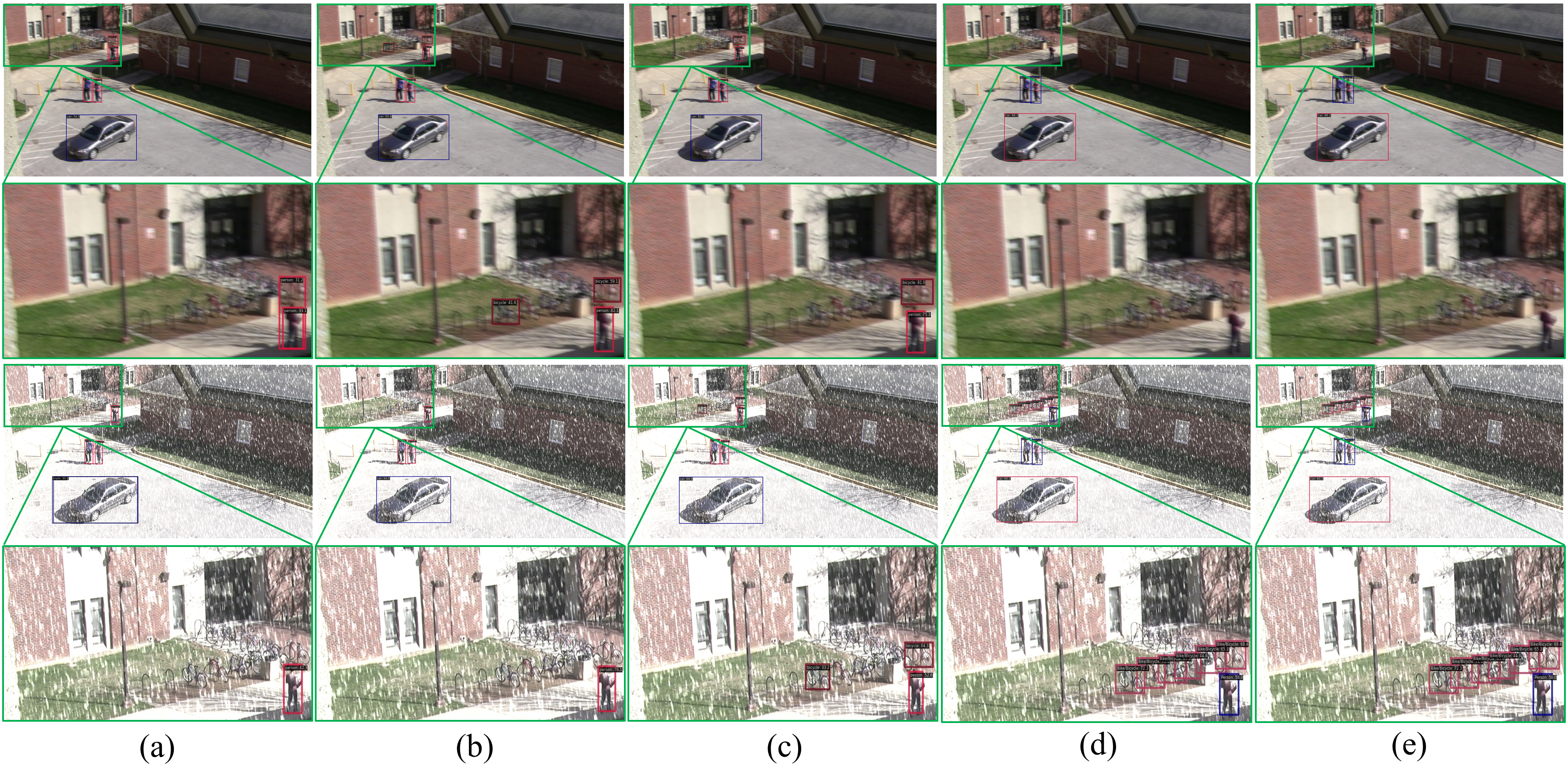}
	\caption{Visual comparative analysis of selected object detection models on test image perturbed with Motion Blur and Snow perturbation (severity level 3). (a) RTMDET, (b) YOLOX, (c) RetinaNet, (d) DETR, and (e) Deformable-DETR. The second and fourth row contains the enlarge regions, depicting misdetection in images of first and third row.}
	\label{fig:visual_results_cmp}
\end{figure*}
\indent Further, to assess the impact of the various image perturbations on the performance of selected object detection models, we tested each model on an image perturbed with Motion Blur and Snow noise with severity level 3. The obtained visual results are shown in Fig.~\ref{fig:visual_results_cmp}. In Fig.~\ref{fig:visual_results_cmp}, the images in second and fourth row with green borders represent the enlarged regions from the detection outputs in first and third row. The enlarged regions depicts the misdetection of objects due to Motion Blur and Snow noise. As can be seen in the Fig.~\ref{fig:visual_results_cmp}, YOLOX and RetinaNet perform well against Motion Blur noise in comparison with RTMDET, DETR, and Deformable-DETR architectures. On the other hand, DETR and Deformable-DETR shows  robustness when tested on image perturbed with Snow noise, as shown in second last and last column of fourth row, Fig.~\ref{fig:visual_results_cmp}.\\
\indent Thus, the experimentation with various image perturbations highlights the varying degrees of robustness among different object detection models. Specifically, YOLOX and RetinaNet demonstrate superior performance against Motion Blur noise, whereas DETR and Deformable-DETR exhibit greater resilience in scenarios affected by Snow noise. These findings underscore the importance of understanding and mitigating the impact of environmental factors on object detection accuracy.
\begin{figure*}[t]
	\centering
	\includegraphics[width=0.87\linewidth]{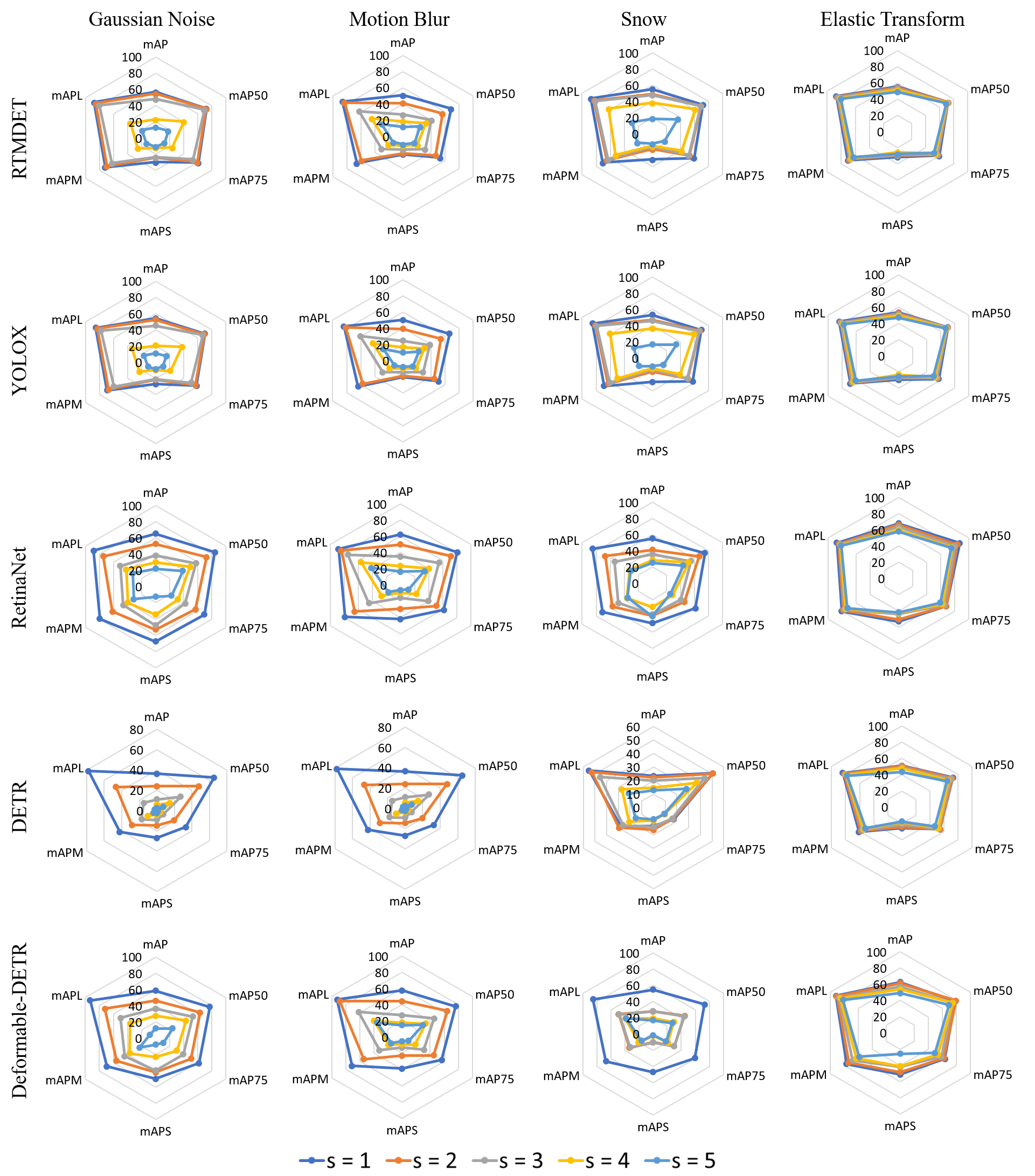}
	\caption{The obtained quantitative results in terms of $mAP$, $mAP_{50}$, $mAP_{75}$, $mAP_{S}$, $mAP_{M}$, and $mAP_{L}$ on test images perturbed with Gaussian Noise, Motion Blur, Snow, and Elastic Transform and five different level of perturbation severity (i.e., s = [1:1:5]). }
	\label{fig:visual_cmp_results}
\end{figure*}
\begin{figure*}[t]
    \centering
    \begin{subfigure}{0.45\textwidth}
        \centering
        \includegraphics[width=\linewidth]{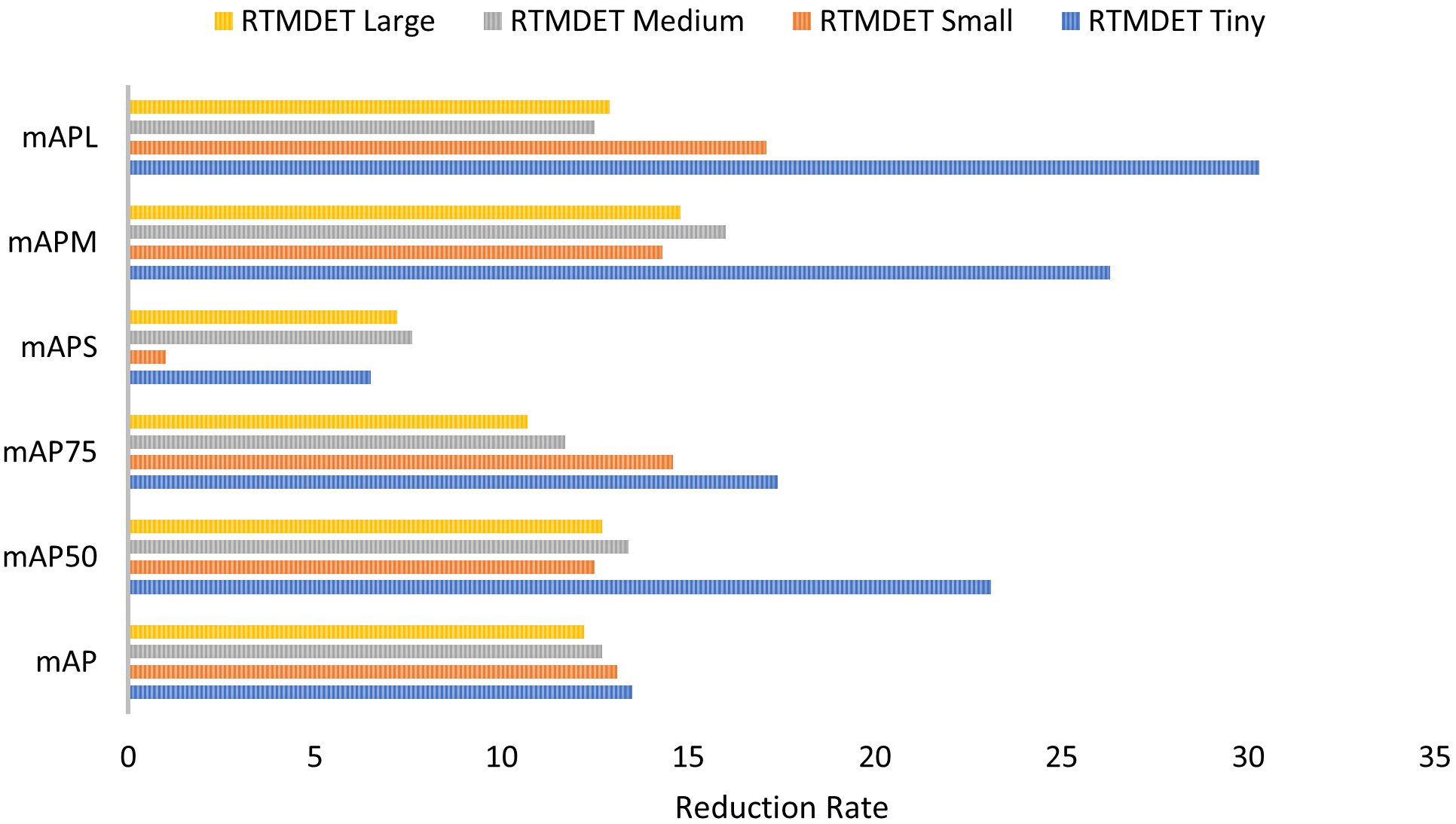}
        \caption{RTMDET}
        \label{fig:sub1}
    \end{subfigure}
    \begin{subfigure}{0.45\textwidth}
        \centering
        \includegraphics[width=\linewidth]{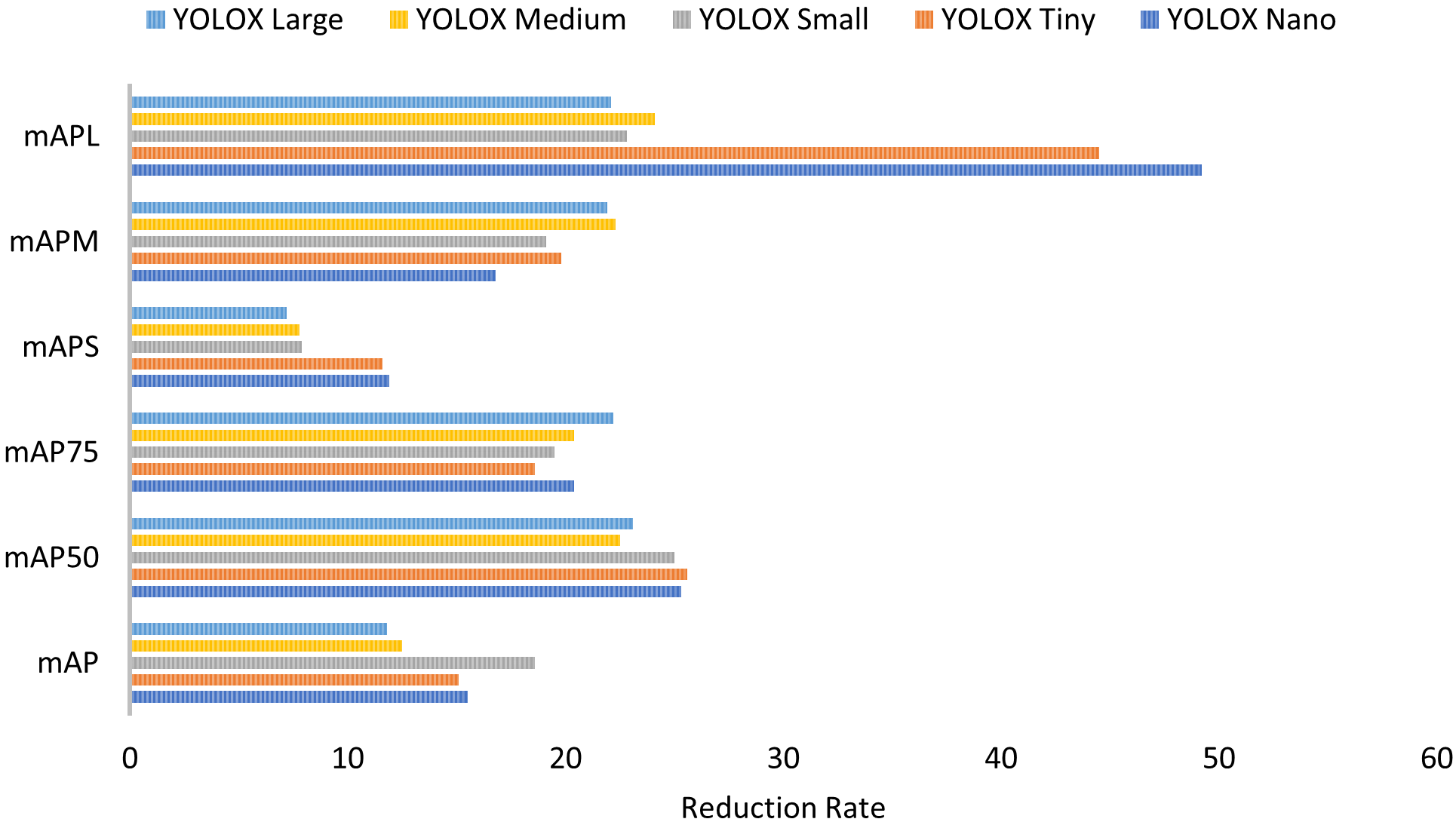}
        \caption{YOLOX}
        \label{fig:sub2}
    \end{subfigure}
    \begin{subfigure}{0.45\textwidth}
        \centering
        \includegraphics[width=\linewidth]{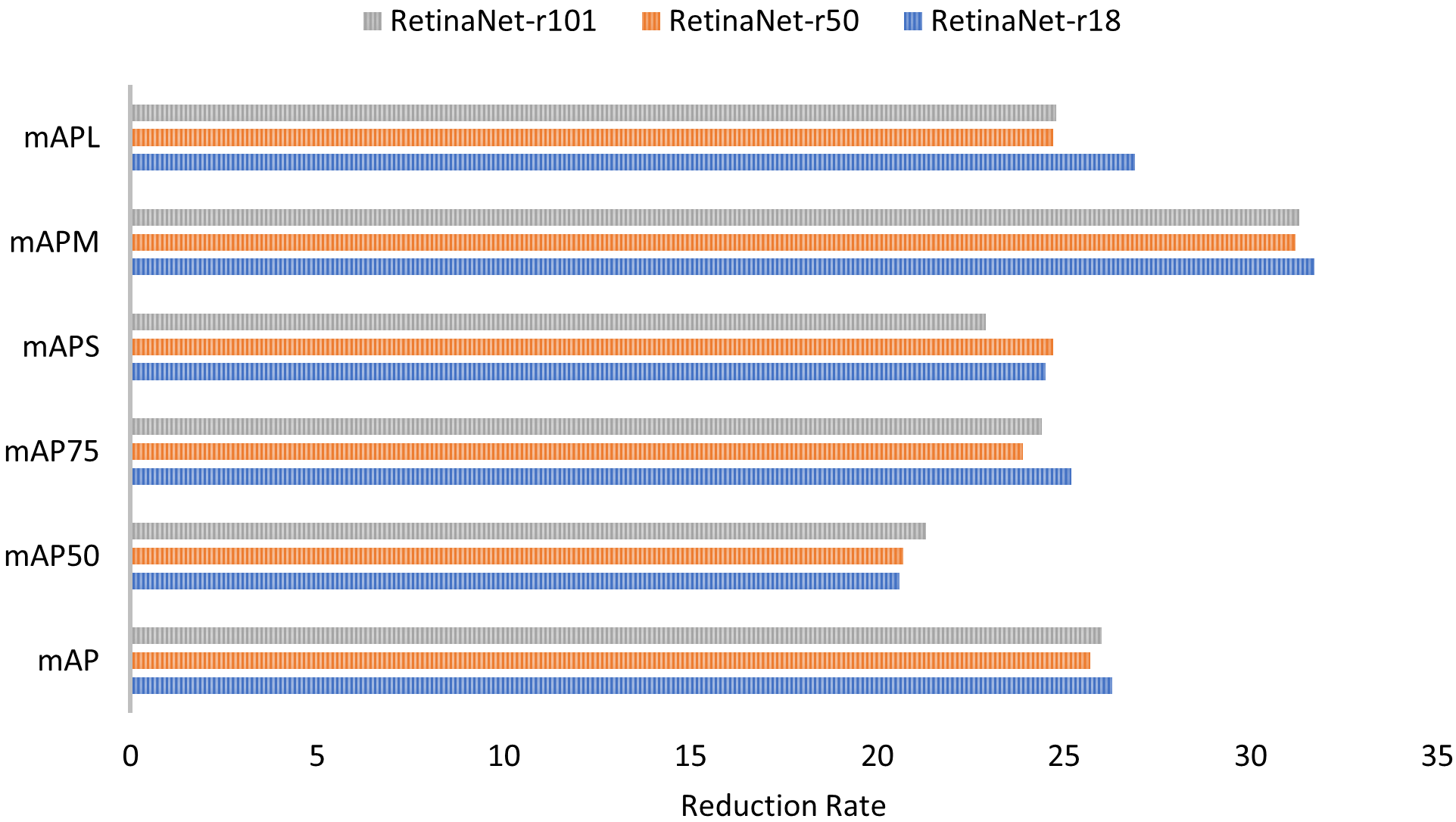}
        \caption{RetinaNet}
        \label{fig:sub3}
    \end{subfigure}
    \begin{subfigure}{0.45\textwidth}
        \centering
        \includegraphics[width=\linewidth]{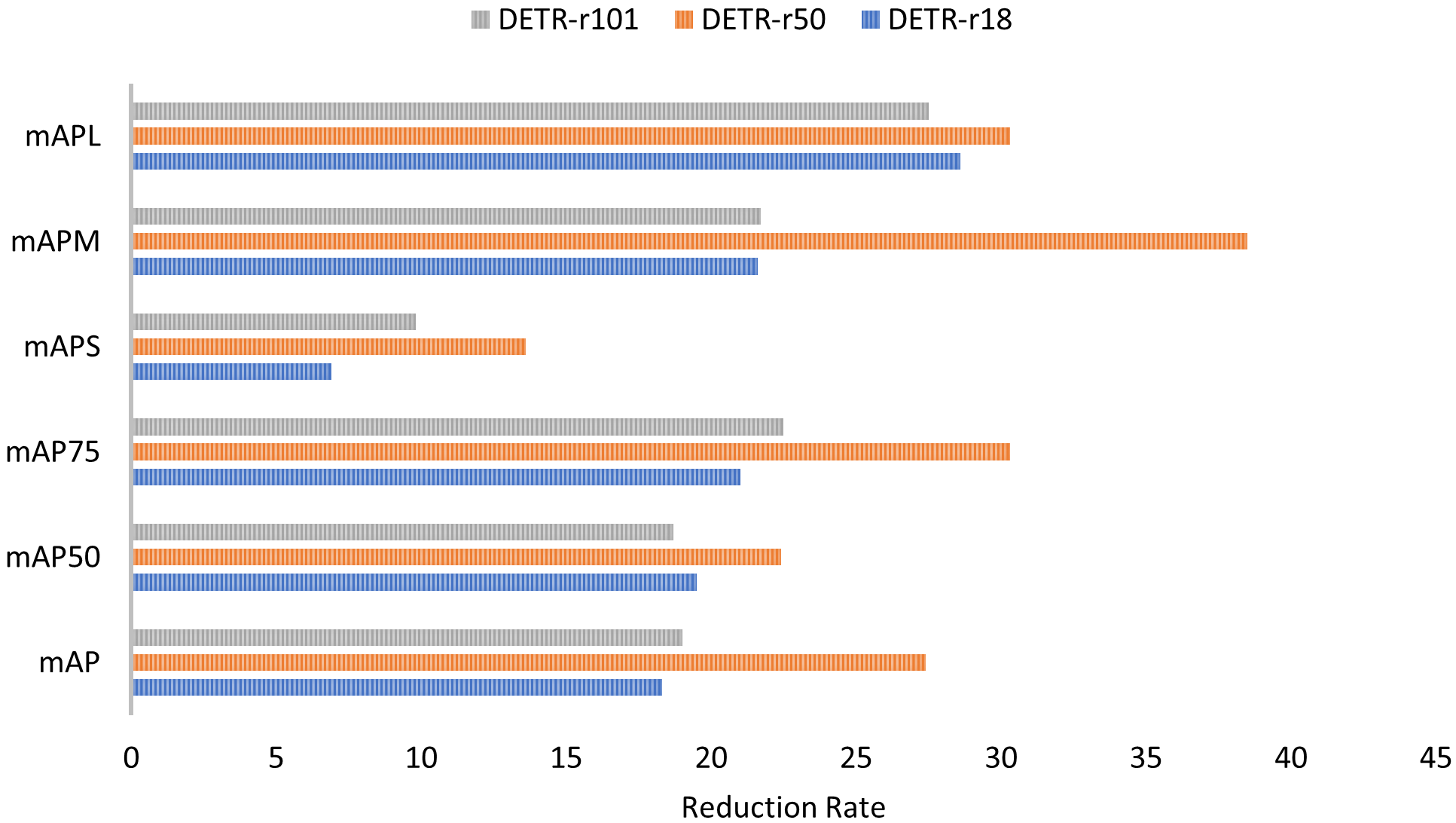}
        \caption{DETR}
        \label{fig:sub4}
    \end{subfigure}
    \begin{subfigure}{0.45\textwidth}
        \centering
        \includegraphics[width=\linewidth]{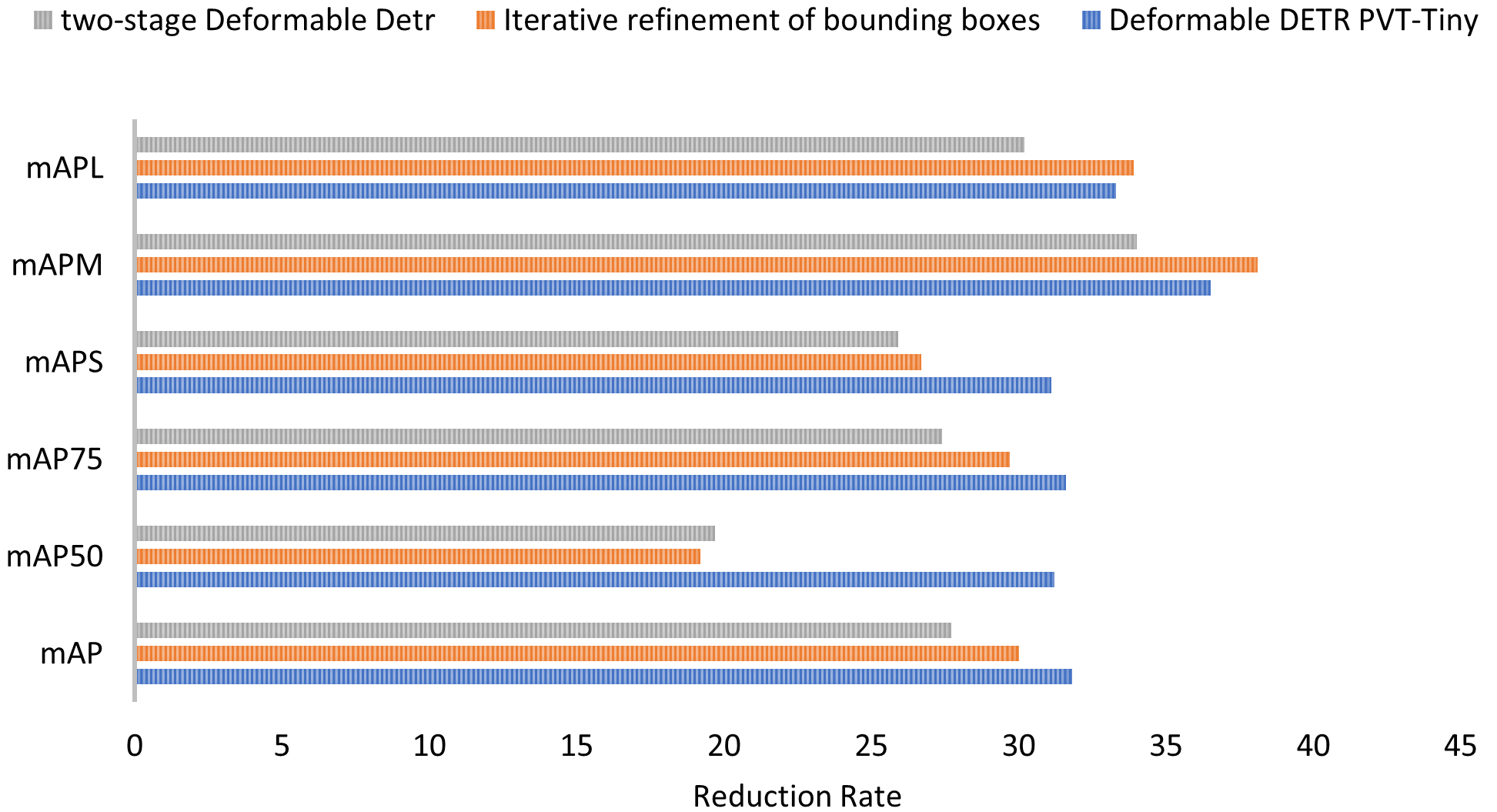}
        \caption{Deformable DETR PVT}
        \label{fig:sub5}
    \end{subfigure}
    \begin{subfigure}{0.45\textwidth}
        \centering
        \includegraphics[width=\linewidth]{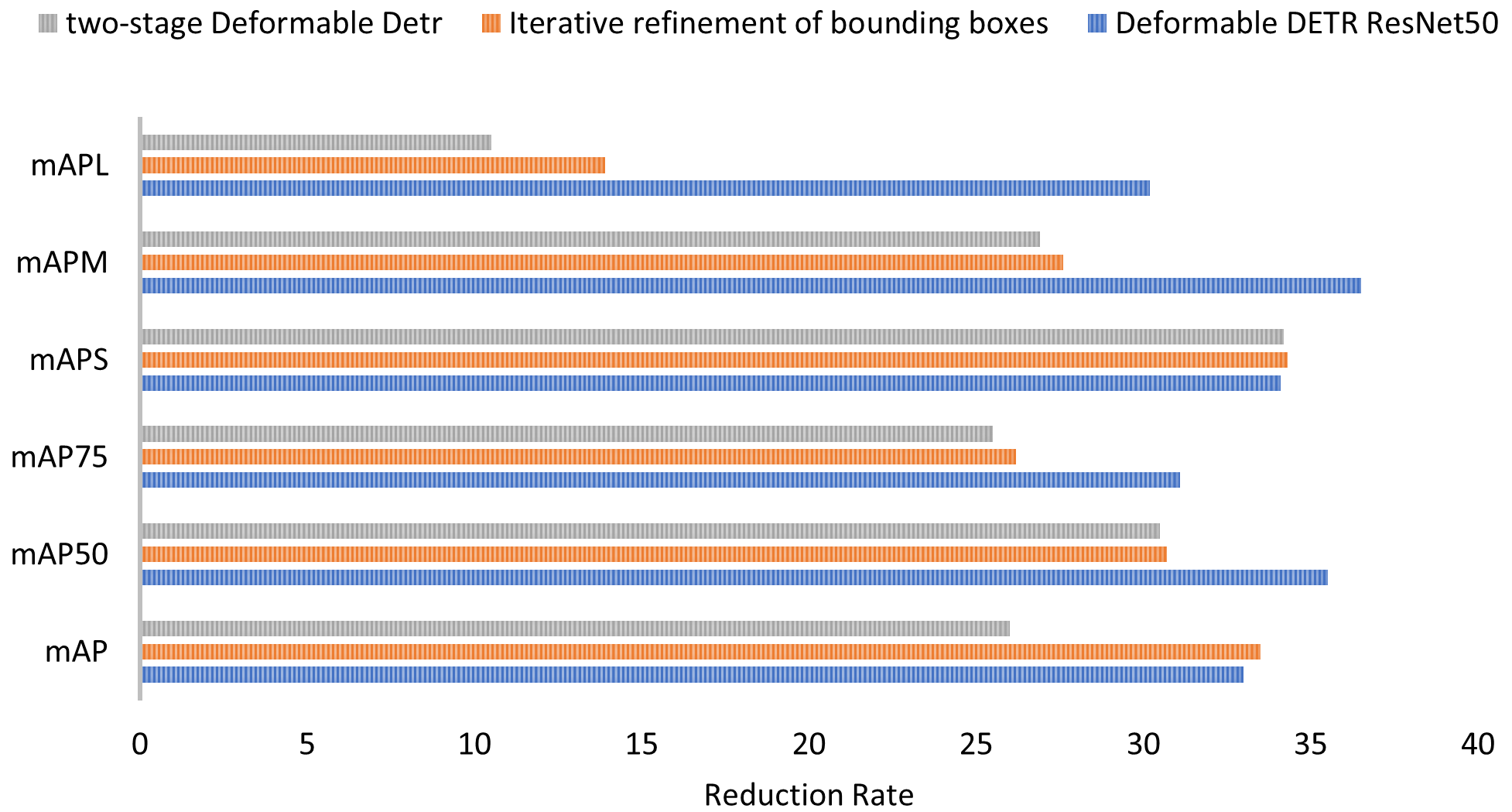}
        \caption{Deformable DETR ResNet50}
        \label{fig:sub6}
    \end{subfigure}
    \caption{Performance reduction of each model on test images perturbed with Gaussian Noise, Motion Blur, Snow, and Elastic  Transform. (a) RTMDET, (b) YOLOX, (c) RetinaNet, (d) DETR, (e) Deformable-DETR with PVT backbone, and (f) Deformable-DETR with ResNet50 backbone.}
    \label{fig:all_reduc}
\end{figure*}

\section{Conclusion}
\label{sec:conclusion}
In this paper, we present two object detection benchmark datasets, OD-VIRAT Large and OD-VIRAT Tiny, comprising realistic surveillance imagery of ten different scenes. The OD-VIRAT Large dataset provides 8.7 million annotated instances 599,996 images, whereas the OD-VIRAT Tiny has 288,901 annotated instances in 19,860 images. The images featured in both benchmarks present challenging characteristics, including complex backgrounds, occluded tiny objects, and objects of varying sizes, which make their detection difficult. To investigate the object detection performance in the presence of these challenging image characteristics, we conducted an extensive experimental evaluation of five state-of-the-art object detection architectures, including RTMDET, YOLOX, RetinaNet, DETR, and Deformable-DETR on the OD-VIRAT Tiny dataset. The obtained quantitative results (in terms of mAP) demonstrate the effectiveness of Deformable-DETR architecture which achieves the best mAP values when trained with ResNet50 backbone. The RetinaNet architecture with ResNet(18,50,101) backbones attains the second-best mAP values across each IoU threshold. Conversely, the performance of RTMDET model is considerably low on OD-VIRAT Tiny test set in comparison with other models. Further, we evaluated the performance of each model included in our benchmarking scheme on images perturbed with different noises/artifacts, including Gaussian Noise, Motion Blur, Snow, and Elastic Transform. We anticipate that the empirical evaluation reported in this paper will establish a standard baseline for future object detection architectures to enhance the performance on challenging surveillance images.\\

\section*{Appendix}
\subsection{Additional Robustness Analysis Results}
This section presents the additional quantitative results obtained from model robustness analysis experiments. To assess the mAP performance of object detection models included in our experimental settings, we evaluated each model on the test set of OD-VIRAT Tiny dataset perturbed with four different type of image noises including Gaussian Noise, Motion Blur, Snow, and Elastic transform. The obtained quantitative results in terms of $mAP$, $mAP_{50}$, $mAP_{75}$, $mAP_{S}$, $mAP_{M}$, and $mAP_{L}$ metrics are depicted in Fig.~\ref{fig:visual_cmp_results}. As it can be notice from the results depicted in Fig.~\ref{fig:visual_cmp_results}, the performance of RTMDET architecture is greatly reduced when tested on a test set perturbed using Motion Blur with five different noise severity levels.The significant reduction can be notice at severity level 5, where the mAP values are reduced approximately 3$\times$ as compare to severity level 1. Similar trend of reduction can be found, when tested YOLOX on a test set perturbed using Gaussian Blur with five different severity levels. The performance of YOLOX model at severity level 4 and 5  is reduced approximately 3$\times$ and 4$\times$, respectively.\\
\indent The performance of RetinaNet is equally reduced by Gaussian Noise and Motion Blur. For instance, when tested RetinaNet architecture on a test set perturbed with Gaussian Noise and Motion Blur, the performance is reduced approximately 3$\times$ at severity level 5 for both noises. Similarly, the performance of DETR architecture is greatly reduced when tested on a test set perturbed using Gaussian Noise and Motion Blur with severity level 5. Surprisingly, the performance of DETR architecture is approximately reduced to less than 10\% for each mAP metric including $mAP$, $mAP_{50}$, $mAP_{75}$, $mAP_{S}$, $mAP_{M}$, and $mAP_{L}$. Finally, the performance of Deformable-DETR architecture is reduced approximately 3$\times$ when tested on a test set perturbed using Gaussian Noise, Motion Blur, and Snow with severity level 5.\\
\indent Besides, we also computed the exact amount of reduction in mAP performance of selected object detection models on a test set perturbed with the above-mentioned image perturbations. It is worth mentioning here, that the average values of each mAP metric including $mAP$, $mAP_{50}$, $mAP_{75}$, $mAP_{S}$, $mAP_{M}$, and $mAP_{L}$ is computed across all perturbations (i.e., Gaussian Noise, Motion Blur, Snow, and Elastic Transform). The obtained reduction values for each model is depicted in Fig.~\ref{fig:all_reduc}.\\

\section*{Acknowledgment}
This research was supported in part by the Air Force Office of Scientific Research (AFOSR) Contract Number FA9550-22-1-0040. The authors would like to acknowledge Dr. Erik Blasch from the Air Force Research Laboratory (AFRL) for his guidance and support on the project. Any opinions, findings, and conclusions or recommendations expressed in this material are those of the author(s) and do not necessarily reflect the views of the Air Force, the Air Force Research Laboratory (AFRL), and/or AFOSR.

\bibliographystyle{IEEEtran}
\bibliography{References}

\begin{thebibliography}{10}
\providecommand{\url}[1]{#1}
\csname url@samestyle\endcsname
\providecommand{\newblock}{\relax}
\providecommand{\bibinfo}[2]{#2}
\providecommand{\BIBentrySTDinterwordspacing}{\spaceskip=0pt\relax}
\providecommand{\BIBentryALTinterwordstretchfactor}{4}
\providecommand{\BIBentryALTinterwordspacing}{\spaceskip=\fontdimen2\font plus
\BIBentryALTinterwordstretchfactor\fontdimen3\font minus \fontdimen4\font\relax}
\providecommand{\BIBforeignlanguage}[2]{{%
\expandafter\ifx\csname l@#1\endcsname\relax
\typeout{** WARNING: IEEEtran.bst: No hyphenation pattern has been}%
\typeout{** loaded for the language `#1'. Using the pattern for}%
\typeout{** the default language instead.}%
\else
\language=\csname l@#1\endcsname
\fi
#2}}
\providecommand{\BIBdecl}{\relax}
\BIBdecl

\bibitem{cheng2022anchor}
G.~Cheng, J.~Wang, K.~Li, X.~Xie, C.~Lang, Y.~Yao, and J.~Han, ``Anchor-free oriented proposal generator for object detection,'' \emph{IEEE Transactions on Geoscience and Remote Sensing}, vol.~60, pp. 1--11, 2022.

\bibitem{chen2023diffusiondet}
S.~Chen, P.~Sun, Y.~Song, and P.~Luo, ``Diffusiondet: Diffusion model for object detection,'' in \emph{Proceedings of the IEEE/CVF International Conference on Computer Vision}, 2023, pp. 19\,830--19\,843.

\bibitem{pu2024rank}
Y.~Pu, W.~Liang, Y.~Hao, Y.~Yuan, Y.~Yang, C.~Zhang, H.~Hu, and G.~Huang, ``Rank-detr for high quality object detection,'' \emph{Advances in Neural Information Processing Systems}, vol.~36, 2024.

\bibitem{roy2022fast}
A.~M. Roy, R.~Bose, and J.~Bhaduri, ``A fast accurate fine-grain object detection model based on yolov4 deep neural network,'' \emph{Neural Computing and Applications}, pp. 1--27, 2022.

\bibitem{li2022dual}
S.~Li, C.~He, R.~Li, and L.~Zhang, ``A dual weighting label assignment scheme for object detection,'' in \emph{Proceedings of the IEEE/CVF Conference on Computer Vision and Pattern Recognition}, 2022, pp. 9387--9396.

\bibitem{li2022yolov6}
C.~Li, L.~Li, H.~Jiang, K.~Weng, Y.~Geng, L.~Li, Z.~Ke, Q.~Li, M.~Cheng, W.~Nie \emph{et~al.}, ``Yolov6: A single-stage object detection framework for industrial applications,'' \emph{arXiv preprint arXiv:2209.02976}, 2022.

\bibitem{wang2023yolov7}
C.-Y. Wang, A.~Bochkovskiy, and H.-Y.~M. Liao, ``Yolov7: Trainable bag-of-freebies sets new state-of-the-art for real-time object detectors,'' in \emph{Proceedings of the IEEE/CVF Conference on Computer Vision and Pattern Recognition}, 2023, pp. 7464--7475.

\bibitem{li2022exploring}
Y.~Li, H.~Mao, R.~Girshick, and K.~He, ``Exploring plain vision transformer backbones for object detection,'' in \emph{European Conference on Computer Vision}.\hskip 1em plus 0.5em minus 0.4em\relax Springer, 2022, pp. 280--296.

\bibitem{lin2014microsoft}
T.-Y. Lin, M.~Maire, S.~Belongie, J.~Hays, P.~Perona, D.~Ramanan, P.~Doll{\'a}r, and C.~L. Zitnick, ``Microsoft coco: Common objects in context,'' in \emph{Computer Vision--ECCV 2014: 13th European Conference, Zurich, Switzerland, September 6-12, 2014, Proceedings, Part V 13}.\hskip 1em plus 0.5em minus 0.4em\relax Springer, 2014, pp. 740--755.

\bibitem{everingham2010pascal}
M.~Everingham, L.~Van~Gool, C.~K. Williams, J.~Winn, and A.~Zisserman, ``The pascal visual object classes (voc) challenge,'' \emph{International journal of computer vision}, vol.~88, pp. 303--338, 2010.

\bibitem{yu2020bdd100k}
F.~Yu, H.~Chen, X.~Wang, W.~Xian, Y.~Chen, F.~Liu, V.~Madhavan, and T.~Darrell, ``Bdd100k: A diverse driving dataset for heterogeneous multitask learning,'' in \emph{Proceedings of the IEEE/CVF conference on computer vision and pattern recognition}, 2020, pp. 2636--2645.

\bibitem{voc2017}
{Everingham, Mark and Van Gool, Luc and Williams, Christopher KI and Winn, John and Zisserman, Andrew}, ``Visual object classes challenge 2012,'' \url{http://host.robots.ox.ac.uk/pascal/VOC/voc2012/}, accessed on Date (May 30, 2024).

\bibitem{coco2017}
{Lin, Tsung-Yi and Maire, Michael and Belongie, Serge and Hays, James and Perona, Pietro and Ramanan, Deva and Doll{\'a}r, Piotr and Zitnick, C Lawrence}, ``Coco: Microsoft common objects in context,'' \url{https://cocodataset.org/}, accessed on Date (May 30, 2024).

\bibitem{shao2019objects365}
S.~Shao, Z.~Li, T.~Zhang, C.~Peng, G.~Yu, X.~Zhang, J.~Li, and J.~Sun, ``Objects365: A large-scale, high-quality dataset for object detection,'' in \emph{Proceedings of the IEEE/CVF international conference on computer vision}, 2019, pp. 8430--8439.

\bibitem{OpenImages}
A.~Kuznetsova, H.~Rom, N.~Alldrin, J.~Uijlings, I.~Krasin, J.~Pont-Tuset, S.~Kamali, S.~Popov, M.~Malloci, A.~Kolesnikov, T.~Duerig, and V.~Ferrari, ``The open images dataset v4: Unified image classification, object detection, and visual relationship detection at scale,'' \emph{IJCV}, 2020.

\bibitem{9573394}
P.~Zhu, L.~Wen, D.~Du, X.~Bian, H.~Fan, Q.~Hu, and H.~Ling, ``Detection and tracking meet drones challenge,'' \emph{IEEE Transactions on Pattern Analysis and Machine Intelligence}, pp. 1--1, 2021.

\bibitem{wang2023v3det}
J.~Wang, P.~Zhang, T.~Chu, Y.~Cao, Y.~Zhou, T.~Wu, B.~Wang, C.~He, and D.~Lin, ``V3det: Vast vocabulary visual detection dataset,'' in \emph{The IEEE International Conference on Computer Vision (ICCV)}, October 2023.

\bibitem{oh2011large}
S.~Oh, A.~Hoogs, A.~Perera, N.~Cuntoor, C.-C. Chen, J.~T. Lee, S.~Mukherjee, J.~Aggarwal, H.~Lee, L.~Davis \emph{et~al.}, ``A large-scale benchmark dataset for event recognition in surveillance video,'' in \emph{CVPR 2011}.\hskip 1em plus 0.5em minus 0.4em\relax IEEE, 2011, pp. 3153--3160.

\bibitem{lyu2022rtmdet}
C.~Lyu, W.~Zhang, H.~Huang, Y.~Zhou, Y.~Wang, Y.~Liu, S.~Zhang, and K.~Chen, ``Rtmdet: An empirical study of designing real-time object detectors,'' \emph{arXiv preprint arXiv:2212.07784}, 2022.

\bibitem{ge2021yolox}
Z.~Ge, S.~Liu, F.~Wang, Z.~Li, and J.~Sun, ``Yolox: Exceeding yolo series in 2021,'' \emph{arXiv preprint arXiv:2107.08430}, 2021.

\bibitem{lin2017focal}
T.-Y. Lin, P.~Goyal, R.~Girshick, K.~He, and P.~Doll{\'a}r, ``Focal loss for dense object detection,'' in \emph{Proceedings of the IEEE international conference on computer vision}, 2017, pp. 2980--2988.

\bibitem{carion2020end}
N.~Carion, F.~Massa, G.~Synnaeve, N.~Usunier, A.~Kirillov, and S.~Zagoruyko, ``End-to-end object detection with transformers,'' in \emph{European conference on computer vision}.\hskip 1em plus 0.5em minus 0.4em\relax Springer, 2020, pp. 213--229.

\bibitem{zhu2020deformable}
X.~Zhu, W.~Su, L.~Lu, B.~Li, X.~Wang, and J.~Dai, ``Deformable detr: Deformable transformers for end-to-end object detection,'' \emph{arXiv preprint arXiv:2010.04159}, 2020.

\bibitem{chen2024cspnext}
X.~Chen, C.~Yang, J.~Mo, Y.~Sun, H.~Karmouni, Y.~Jiang, and Z.~Zheng, ``Cspnext: A new efficient token hybrid backbone,'' \emph{Engineering Applications of Artificial Intelligence}, vol. 132, p. 107886, 2024.

\bibitem{lin2017feature}
T.-Y. Lin, P.~Doll{\'a}r, R.~Girshick, K.~He, B.~Hariharan, and S.~Belongie, ``Feature pyramid networks for object detection,'' in \emph{Proceedings of the IEEE conference on computer vision and pattern recognition}, 2017, pp. 2117--2125.

\bibitem{wang2020cspnet}
C.-Y. Wang, H.-Y.~M. Liao, Y.-H. Wu, P.-Y. Chen, J.-W. Hsieh, and I.-H. Yeh, ``Cspnet: A new backbone that can enhance learning capability of cnn,'' in \emph{Proceedings of the IEEE/CVF conference on computer vision and pattern recognition workshops}, 2020, pp. 390--391.

\bibitem{he2016deep}
K.~He, X.~Zhang, S.~Ren, and J.~Sun, ``Deep residual learning for image recognition,'' in \emph{Proceedings of the IEEE conference on computer vision and pattern recognition}, 2016, pp. 770--778.

\bibitem{wang2021pyramid}
W.~Wang, E.~Xie, X.~Li, D.-P. Fan, K.~Song, D.~Liang, T.~Lu, P.~Luo, and L.~Shao, ``Pyramid vision transformer: A versatile backbone for dense prediction without convolutions,'' in \emph{Proceedings of the IEEE/CVF international conference on computer vision}, 2021, pp. 568--578.

\bibitem{mmdetection}
K.~Chen, J.~Wang, J.~Pang, Y.~Cao, Y.~Xiong, X.~Li, S.~Sun, W.~Feng, Z.~Liu, J.~Xu, Z.~Zhang, D.~Cheng, C.~Zhu, T.~Cheng, Q.~Zhao, B.~Li, X.~Lu, R.~Zhu, Y.~Wu, J.~Dai, J.~Wang, J.~Shi, W.~Ouyang, C.~C. Loy, and D.~Lin, ``{MMDetection}: Open mmlab detection toolbox and benchmark,'' \emph{arXiv preprint arXiv:1906.07155}, 2019.

\end{thebibliography}

\end{document}